\documentclass[letterpaper, 10 pt, conference]{ieeeconf}  % Comment this line out if you need a4paper

\IEEEoverridecommandlockouts                              % This command is only needed if 
\usepackage{amssymb}
\usepackage{amsmath}
\usepackage{graphicx}
\usepackage{booktabs}
\usepackage{array}
\usepackage{changepage}
\usepackage{mathtools}
\usepackage{svg}
\usepackage{bbm}
\usepackage{wrapfig}
\usepackage{subcaption}
\usepackage{adjustbox}
\usepackage{algorithm}
\usepackage{algpseudocode}
\usepackage{algorithmicx}
\usepackage{hyphenat}
\usepackage{multirow}
\usepackage{makecell}
\usepackage{titlesec}
\usepackage{hyperref}

\titlespacing*{\section}{0pt}{0.5ex plus 1ex minus .2ex}{0.3ex plus .2ex}

\usepackage{float}

\usepackage{balance} % For better column balancing
\usepackage{dblfloatfix} % For better figure placement in double column

\usepackage{xcolor}
\definecolor{orange}{rgb}{1,0.5,0}
\definecolor{internationalorange}{rgb}{1.0, 0.31, 0.0}

\newcommand{\xxnote}[3]{}
\ifx\hidenotes\undefined
  \usepackage{color}
  \renewcommand{\xxnote}[3]{\color{#2}{#1: #3}}
\fi

{\begin{list}{}%
  {\setlength{\leftmargin}{#1}}%
  \item[] \itshape}
{\end{list}}
\usepackage{xspace}
\newcommand\algname[2]{\texttt{#1}{#2}\xspace}

\newcommand{\gOOD}{\algname{RISE}{}}

\newtheorem{assumption}{Assumption}

\newcommand\taskname[2]{\texttt{#1}{#2}\xspace}

\newcommand\mug{\taskname{mug-cleanup}{}}
\newcommand\squarenut{\taskname{square-peg}{}}
\newcommand\squarehook{\taskname{square-hook}{}}
\newcommand\twopiece{\taskname{piece-assembly}{}}

\newcommand\threading{\taskname{threading}{}}

\newcommand\lamp{\taskname{lampshade}{}}
\newcommand\oneleg{\taskname{one-leg}{}}
\newcommand\cloth{\taskname{cloth folding}{}}

\title{Using Non-Expert Data to Robustify Imitation Learning via Offline Reinforcement Learning}

\author{
    Kevin Huang\thanks{*Equal contribution}$^*$$^{1}$,
    Rosario Scalise$^*$$^{1}$,
    Cleah Winston$^{1}$,
    Ayush Agrawal$^{1}$,
    Yunchu Zhang$^{1}$,\\
    Rohan Baijal$^{1}$,
    Markus Grotz$^{1}$,
    Byron Boots$^{1}$,
    Benjamin Burchfiel$^{2}$,
    Masha Itkina$^{2}$,
    Paarth Shah$^{2}$,
    and
    Abhishek Gupta$^{1}$
    \thanks{$^{1}$ University of Washington}
    \thanks{$^{2}$ Toyota Research Institute (TRI)}
    \thanks{Correspondence to \{kehuang, rosario\}@cs.washington.edu}
}

\begin{document}

\maketitle
\thispagestyle{empty}
\pagestyle{empty}

%%%%%%%%%%%%%%%%%%%%%%%%%%%%%%%%%%%%%%%%%%%%%%%%%%%%%%%%%%%%%%%%%%%%%%%%%%%%%%%%
\begin{abstract}

Imitation learning has proven effective for training robots to perform complex tasks from expert human demonstrations. However, it remains limited by its reliance on high-quality, task-specific data, restricting adaptability to the diverse range of real-world object configurations and scenarios. In contrast, non-expert data---such as play data, suboptimal demonstrations, partial task completions, or rollouts from suboptimal policies---can offer broader coverage and lower collection costs. However, conventional imitation learning approaches fail to utilize this data effectively. To address these challenges, we posit that with right design decisions, offline reinforcement learning can be used as a tool to harness non-expert data to enhance the performance of imitation learning policies. We show that while standard offline RL approaches can be ineffective at actually leveraging non-expert data under the sparse data coverage settings typically encountered in the real world, simple algorithmic modifications can allow for the utilization of this data, without significant additional assumptions. Our approach shows that broadening the support of the policy distribution can allow imitation algorithms augmented by offline RL to solve tasks robustly, showing considerably enhanced recovery and generalization behavior. In manipulation tasks, these innovations significantly increase the range of initial conditions where learned policies are successful when non-expert data is incorporated. Moreover, we show that these methods are able to leverage \emph{all} collected data, including partial or suboptimal demonstrations, to bolster task-directed policy performance. This underscores the importance of algorithmic techniques for using non-expert data for robust policy learning in robotics. Paper website: \url{https://uwrobotlearning.github.io/RISE-offline/}
\end{abstract}

%%%%%%%%%%%%%%%%%%%%%%%%%%%%%%%%%%%%%%%%%%%%%%%%%%%%%%%%%%%%%%%%%%%%%%%%%%%%%%%%
\section{Introduction}
Imitation learning has enabled remarkable progress in robot learning, training reactive closed-loop policies from high-quality demonstrations. These methods typically perform supervised learning using expressive policy classes such as diffusion models parameterized with large neural networks~\cite{chi23diffusion, zhao23act, zare2023surveyimitationlearningalgorithms}. Despite impressive performance under conditions similar to the training distribution, these policies can be quite brittle beyond this setting. They show vulnerability to out-of-distribution (OOD) scenarios, where even minor deviations in object configurations or environmental conditions can lead to failure~\cite{majumdar25redteaming}. A natural way to address policy fragility is to simply collect more expert data, broadening the coverage of expert demonstrations. The challenge is that collecting such data can be expensive and scale poorly, requiring an impractical amount of data collection to cover combinatorial scene conditions. As a result, policies trained with standard imitation learning can struggle to handle real-world variability on deployment.

The formulation of imitation learning through supervised learning requires (near) optimal task demonstrations, which have to be carefully curated per task and can be difficult to collect at scale, especially for high precision or high dexterity problems. In most large-scale data collection efforts, not all data satisfies these criteria. This results in a significant fraction of collected data being discarded through the process of data curation/filtering ~\cite{belkhale23data, hejna25curation, agia25cupid}, despite this data containing useful information about the dynamics of the world. This begs the question - \emph{can cheaper sources of non-optimal data beyond expert, task-specific demonstrations be used to improve the performance of imitation learning?} In this work, we study how cheaper and typically more abundant data sources such as undirected play data, unsuccessful/partial demonstrations (whether from a human demonstrator or policy rollouts), or data from other tasks can be made useful for improving the robustness of imitation learning. 

\begin{figure}[!t]
    \centering
    \includegraphics[width=\columnwidth]{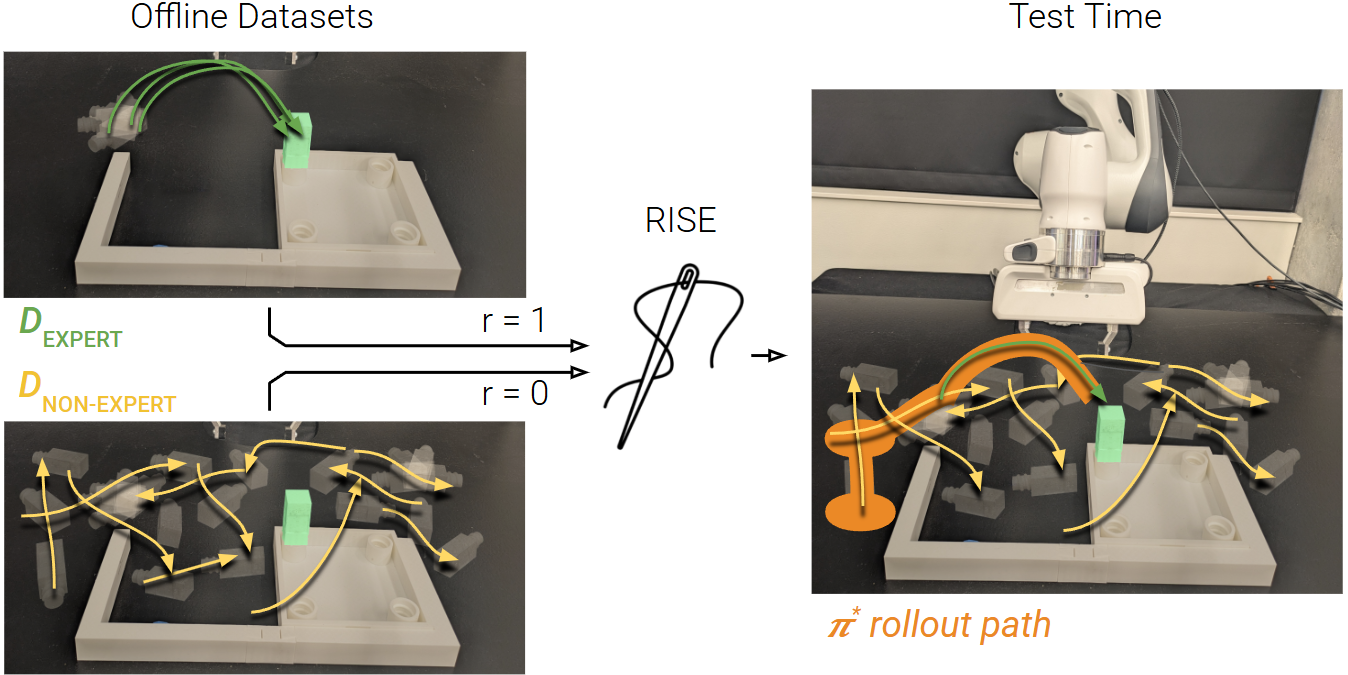}
    \caption{\footnotesize{\gOOD enables non-expert data to be stitched with alongside expert data to provide robust, high-coverage behavior for robotic manipulation. This allows for \emph{all collected data} to be used to allow the policy to recover from novel OOD states.}}
    \label{fig:teaser}
    \vspace{-10pt}
\end{figure}

We study this problem through the lens of offline reinforcement learning (RL). Typical offline RL algorithms ~\cite{cql, brac, idql} treat all data (whether it be optimal or suboptimal) as arbitrary off-policy data and synthesize optimal policies through reward-based, temporal-difference learning~\cite{sutton2018reinforcement}. Directly applying standard offline RL methods for leveraging suboptimal data can become challenging for real-world problems, since reward can be difficult to specify without considerable domain knowledge or privileged information. In this work, we propose an alternative, instantiating an offline RL algorithm that can learn from simple binary rewards, $1$ for optimal data and $0$ for suboptimal data, which collected demonstrations often naturally categorize into. This allows the learning algorithm to make use of the suboptimal data to learn how to \emph{recover} the system back to states in the expert state distribution, while replicating optimal behavior on these expert states. This naturally robustifies the policy to solve tasks from a diversity of states beyond a narrow range of expert states, while requiring minimal infrastructural modifications and assumptions beyond that of typical imitation learning. 

While offline RL via dynamic programming can in principle ``stitch'' together useful segments from suboptimal data for data-efficient recovery, we find that practical instantiations of offline RL methods in high-dimensional state-action spaces can fail to demonstrate this stitching capability without an impractically high degree of data coverage. To address the challenges resulting from the lack of data coverage, we introduce a notion of ``fuzziness'' into the state representation. Specifically, we enforcing a notion of local smoothness on the policy via Lipschitz continuity. For recoverable OOD states, doing so significantly improves the policy’s ability to ``stitch'' offline data. This enables suboptimal data to easily be used for improving the robustness of imitation learning, even in low data coverage regimes. 

We make the following contributions - 1) we introduce an offline RL framework for leveraging non-expert data to robustify imitation learning policies, 2) we show the pitfalls of standard offline RL in the low data regime, and introduce the Robust Imitation by Stitching from Experts (\gOOD) algorithm, to improve trajectory stitching 3) We show that \gOOD is effective across various types of non-optimal data - ranging from undirected play data to suboptimal demonstrations or policy evaluation rollouts, and even multitask datasets, 4) We demonstrate the efficacy of \gOOD on various tabletop manipulation tasks in simulation and furniture assembly tasks on a real robot.
\section{Related Work}

\textbf{Imitation Learning:} Imitation learning methods aim to learn closed loop policies from near optimal demonstration data. This is a well studied field, with a plethora of work~\cite{hussein2017imitation, osa2018algorithmic, argall2009survey, ravichandar2020recent} on methods and applications. Work in imitation learning has primarily focused on either dealing with compounding error \cite{dagger, ke2024ccilcontinuitybaseddataaugmentation, fimitation, diffdagger, laskey2017dartnoiseinjectionrobust}, incorporating richer generative distributions \cite{chi23diffusion, bet, alohaunleashed} or using robust policy backbones~\cite{black2024pi0visionlanguageactionflowmodel, rt2, openvla, karamcheti2023voltron}. In this work, we show that in addition to expert demonstrations, \emph{non-expert} data can be leveraged to robustify imitation learning. 

\textbf{Offline Reinforcement Learning:} Offline reinforcement learning is a closely related subarea of research, where pre-collected off-policy datasets are used to synthesize task-directed behavior \cite{levine2020offlinereinforcementlearningtutorial}. These methods do not typically assume that pre-collected data is optimal, instead using rewards to infer which behaviors are optimal from offline datasets. Offline RL methods come in many forms - importance sampling-based ~\cite{jiang16doubly, fang24diffusion}, model-based ~\cite{yu20mopo, kidambi20morel}, dynamic-programming-based \cite{kostrikov2021offlinereinforcementlearningimplicit, brac, cql, idql}. Many of these methods operate on the principle of \emph{pessimism} - assuming the worst outside of the training distribution. This restricts the synthesized behavior to compositions of behaviors within the training distribution, often referred to as ``stitching". Importantly, most of these methods still rely on access to rewards at training, an often onerous assumption that makes these methods difficult to use. Perhaps most relevant to this work is SQIL~\cite{reddy2019sqilimitationlearningreinforcement}, which performs offline RL on a mixture of optimal demonstration data and suboptimal data. Our findings indicate that in sparse-coverage problems, SQIL can be insufficient for data stitching, and thus, we propose an alternative that allows for better transitions from non-expert data. 

% Recovery 
\textbf{Out-of-distribution Recovery:} A set of prior methods have considered techniques for recovery back to the manifold of expert behavior, so as to robustify learned policy behavior~\cite{gao2025outofdistributionrecoveryobjectcentrickeypoint, diffdagger, ke2024ccilcontinuitybaseddataaugmentation, backtomanifold}. Prior work~\cite{gao2025outofdistributionrecoveryobjectcentrickeypoint} aims to use keypoint driven gradients to recover to the training distribution, using explicit pose and keypoint estimation and an inverse controller. \cite{backtomanifold} uses equivariance to learn a recovery controller back to the expert manifold. In contrast, \gOOD does not rely on explicit object and state representations, and does not have to learn a separate policy and recovery controller. Prior work does local recovery using synthetic data, via generative models \cite{diffdagger} or learned dynamics \cite{ke2024ccilcontinuitybaseddataaugmentation}. Since these models are only valid in local regions around the data, they struggle with global notions of recovery, as is enabled by \gOOD. Perhaps most relevant is \cite{ilid}, which identifies sub-trajectories in suboptimal data that recover to expert states and selectively adds these to imitation learning. We show that \gOOD is significantly more performant and data efficient than \cite{ilid} due to the ability to stitch trajectories. 
\section{Background}
\label{sec:background}

\begin{figure*}[!t]
    \centering
    \includegraphics[width=0.9\textwidth]{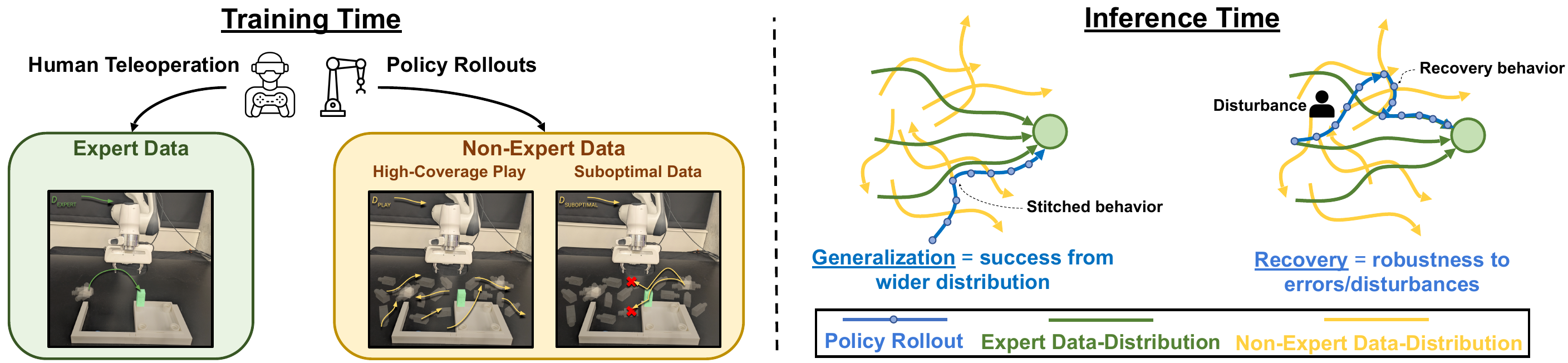}
    \caption{\footnotesize{Various types of data on the \oneleg task is shown: expert, high-coverage, and suboptimal, which can be collected by a human or from autonomous policy rollouts (for example during evaluation). \gOOD is able to use combinations of different expert and non-expert datasets to improve policy robustness. By stitching trajectories from non-expert data, \gOOD policies can succeed from much wider distributions and are robust to disturbances.}}
    \label{fig:recoveryfig}
    \vspace{-15pt}
\end{figure*}

\textbf{Imitation Learning:} We consider an episodic finite-horizon MDP given by $\mathcal{M} = \left\{ \mathcal{S}, \mathcal{A}, p, r, \gamma, H \right\}$, with standard notation~\cite{sutton2018reinforcement}. A policy $\pi$ is a function that maps $s \in \mathcal{S}$ to a distribution over $a \in \mathcal{A}$, and its optimality can be measured by $J(\pi) \coloneqq \mathbb{E}\left [ \sum_{t=0}^{H} \gamma^t r(s_t, a_t) | s_0 \sim p_0, a_t \sim \pi( \cdot | s_t) \right ]$. In the imitation learning problem, we are given a set of demonstrations $\mathcal{D}_E = \{(s_j, a_j)\}_j$ generated from rolling out a (near) expert policy $\pi_E$, from an initial state distribution $p_0$. Given this data, behavior cloning methods learn a policy $\hat{\pi_E}$ via a supervised learning objective: $\hat{\pi_\theta} \leftarrow \max_{\theta} \mathbb{E}_{(s,a) \sim \mathcal{D}_E} \left[ \log (\pi_\theta(a | s)) \right]$. While we parameterize $\pi$ as a conditional diffusion model~\cite{chi23diffusion}, our formulation is equally applicable to $\pi$ being any expressive generative model~\cite{bet, chi23diffusion, parrot}. While $\pi_\theta$ is performant for ``in-distribution" initial conditions $s_0 \sim p_0(\cdot)$, it can be suboptimal when evaluated from OOD conditions $s_0~\sim~p_{\text{test}}(\cdot)$.

% While the behavior cloning objective is typically applied to data from a near-optimal policy, we aim to utilize non-expert data to improve policy performance. 

\textbf{Offline Reinforcement Learning:} Offline RL learns optimal policies from a fixed offline (potentially suboptimal) dataset of transitions $\mathcal{D} = \{(s, a, s', r)_i\}_{i=1}^N$, without requiring online data collection as is common in RL. Offline RL assumes access to labeled rewards $r$, finding a reward-maximizing policy within the support of the offline data. In offline RL literature~\cite{levine20orl}, a majority adopt the mechanism of off-policy RL with an additional element of ``conservatism" to avoid propagating counterfactual OOD value estimates.

We specifically build on a popular, yet simple offline RL variant -- Implicit Diffusion Q-Learning (IDQL)~\cite{idql}, that avoids explicitly imposing conservatism by constraining the policy~\cite{brac} or regularizing the critic~\cite{cql}. Instead, this work proposes to be conservative by approximating an expectile $\tau$ within the distribution of actions, thereby \emph{implicitly} implementing the principle of conservatism. IDQL first learns a parameterized Q-function $Q_\phi(s, a)$ and value function $V_{\psi}$ using the following objective:

\vspace{-5pt}
\begingroup\small
\begin{align} \label{eq:iql_objective}
    \mathcal{L}_V(\psi) &= \mathbb{E}_{s, a \sim \mathcal{D}} \left[ L^{\tau}_2 (Q_\phi (s, a) - V_{\psi}(s)) \right] \\
    \mathcal{L}_Q(\phi) &= \mathbb{E}_{(s, a, s') \sim \mathcal{D}} \left[ \left( r(s, a) + \gamma V_{\psi}(s') - Q_{\phi}(s, a)\right)^2 \right] 
\end{align} 
\endgroup

\noindent where $L_2^{\tau}(x) = \left| \tau - \mathbbm{1}(x < 0) \right|x^2$ leads to learning of the $\tau$ expectile of the action distribution. Given the Q-value function $Q_\phi (s, a)$, IDQL then extracts the optimal policy $\pi^*(a|s)$ through simple non-parametric test-time optimization -- $\pi^*(a|s) = \underset{a \in \{a_1,\dots, a_K\} \sim \pi_B(a|s)}{\text{argmax}} Q_\phi(s, a)$. Samples are drawn from $\pi_B(a|s)$, the ``behavior policy", representing the estimated marginal state-conditional distribution of actions in the training data. The behavior policy $\pi_B(a|s)$ can be obtained through any standard maximum likelihood (or similar) procedure on the offline data, in this case using a diffusion modeling objective~\cite{hodiffusion, chi23diffusion}. 
\section{\gOOD: Leveraging Suboptimal Data for Robust Imitation Learning}
\label{sec:method}

We begin by describing the problem setting (Section~\ref{sec:problem}), followed by an instantiation of a solution technique using offline RL (Section~\ref{subsec:learning_from_non_expert}). We then describe the pitfalls of offline RL methods in low-data regimes, and propose simple algorithmic solutions to these challenges (Section~\ref{sec:smoothness}). 

\subsection{Setting: Robustifying Policies with Non-Expert Data} 
\label{sec:problem}
We will assume access to a dataset of expert state-action tuples $\mathcal{D}_{\text{E}} = \{(s_i, a_i)\}_{i=1}^N$ drawn from an expert, $\pi_E$. This is augmented with a dataset of potentially non-expert state-action tuples $\mathcal{D}_{\text{NE}} = \{(s_i, a_i)\}_{i=1}^M$, where $N \ll M$. The goal is to devise a learning procedure that synthesizes a policy from $\mathcal{D}_{\text{E}}$ and $\mathcal{D}_{\text{NE}}$ that maximizes the task performance across a range of initial conditions. Note that the agent is \textbf{not} provided with labeled rewards $r$ (as is typical in offline RL) during training, only receiving labels of whether the offline data belongs to the expert dataset $\mathcal{D}_{\text{E}}$ or the non-expert dataset $\mathcal{D}_{\text{NE}}$. While non-expert data $\mathcal{D}_{\text{NE}}$ can take many forms, of particular interest are high-coverage datasets, such as undirected ``play" data, multi-task data or closed-loop rollout data collected during evaluations. Partial demonstrations or failures can also provide information about the dynamics of the environment despite being unsuitable for direct imitation. We aim to instantiate a simple, scalable algorithm to augment imitation learning to be able to make use of this non-expert data $\mathcal{D}_{\text{NE}}$. 

\subsection{Learning from Non-Expert Data without Explicit Reward Annotations} \label{subsec:learning_from_non_expert}

While the expert dataset $\mathcal{D}_{\text{E}}$ can simply be copied via typical behavior cloning~\cite{chi23diffusion}, it is not as clear how to use $\mathcal{D}_{\text{NE}}$. We make a simple insight in this work -- while non-expert data $\mathcal{D}_{\text{NE}}$ may not capture expert behavior directly, it conveys information about the dynamics of the environment. This allows a robotic agent to \emph{recover} from an OOD state beyond the expert distribution back to the distribution of expert states in $\mathcal{D}_{\text{E}}$ (Fig~\ref{fig:recoveryfig}), from which the expert can reliably succeed. 

\begin{figure*}[!htb]
    \centering
    % Use full text width for better spacing across both columns
    \begin{subfigure}[b]{0.29\textwidth}
        \centering
        \includegraphics[width=\textwidth]{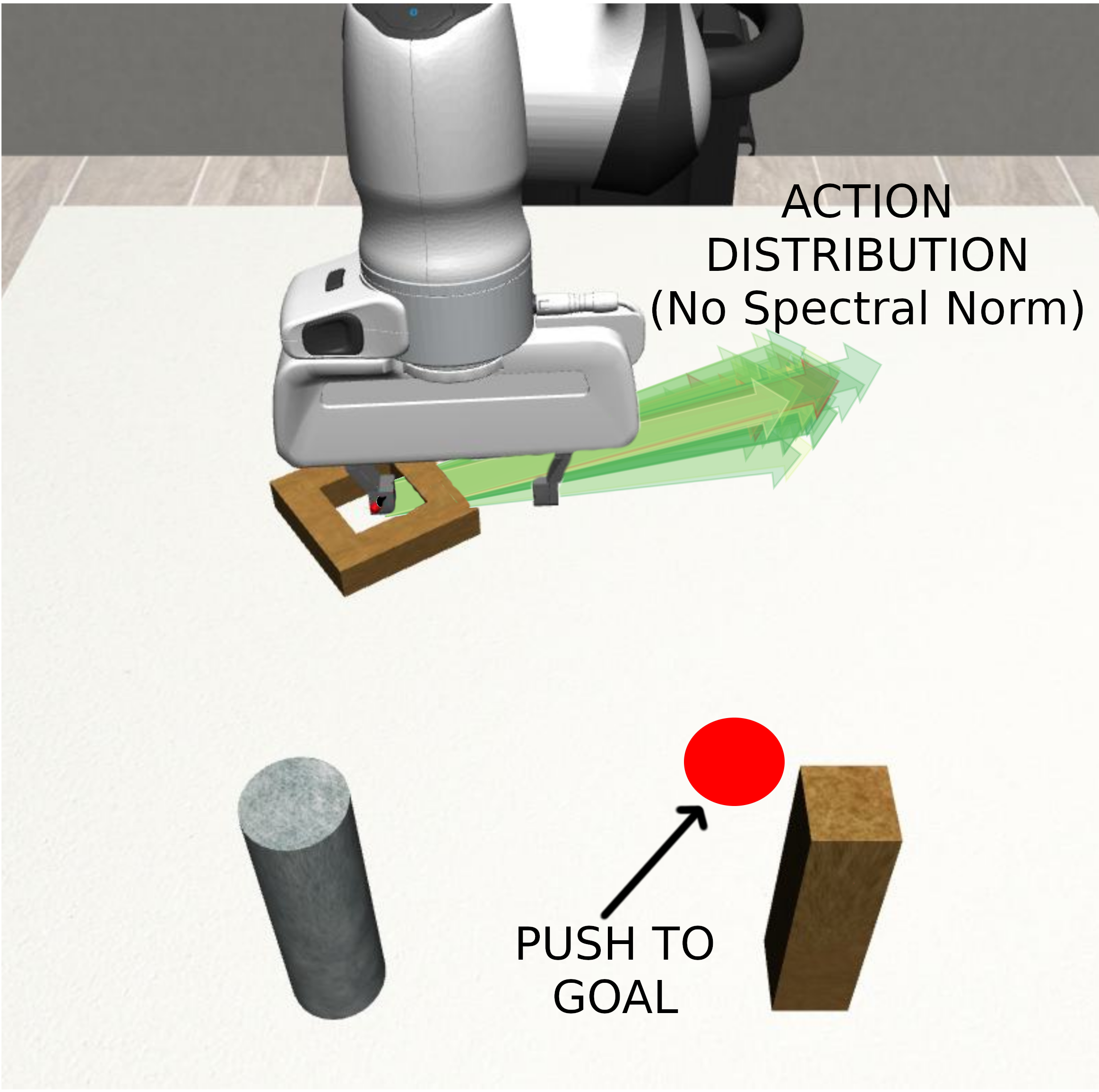}
        \caption{}
        \label{fig:img1}
    \end{subfigure}\hfill%
    \begin{subfigure}[b]{0.29\textwidth}
        \centering
        \includegraphics[width=\textwidth]{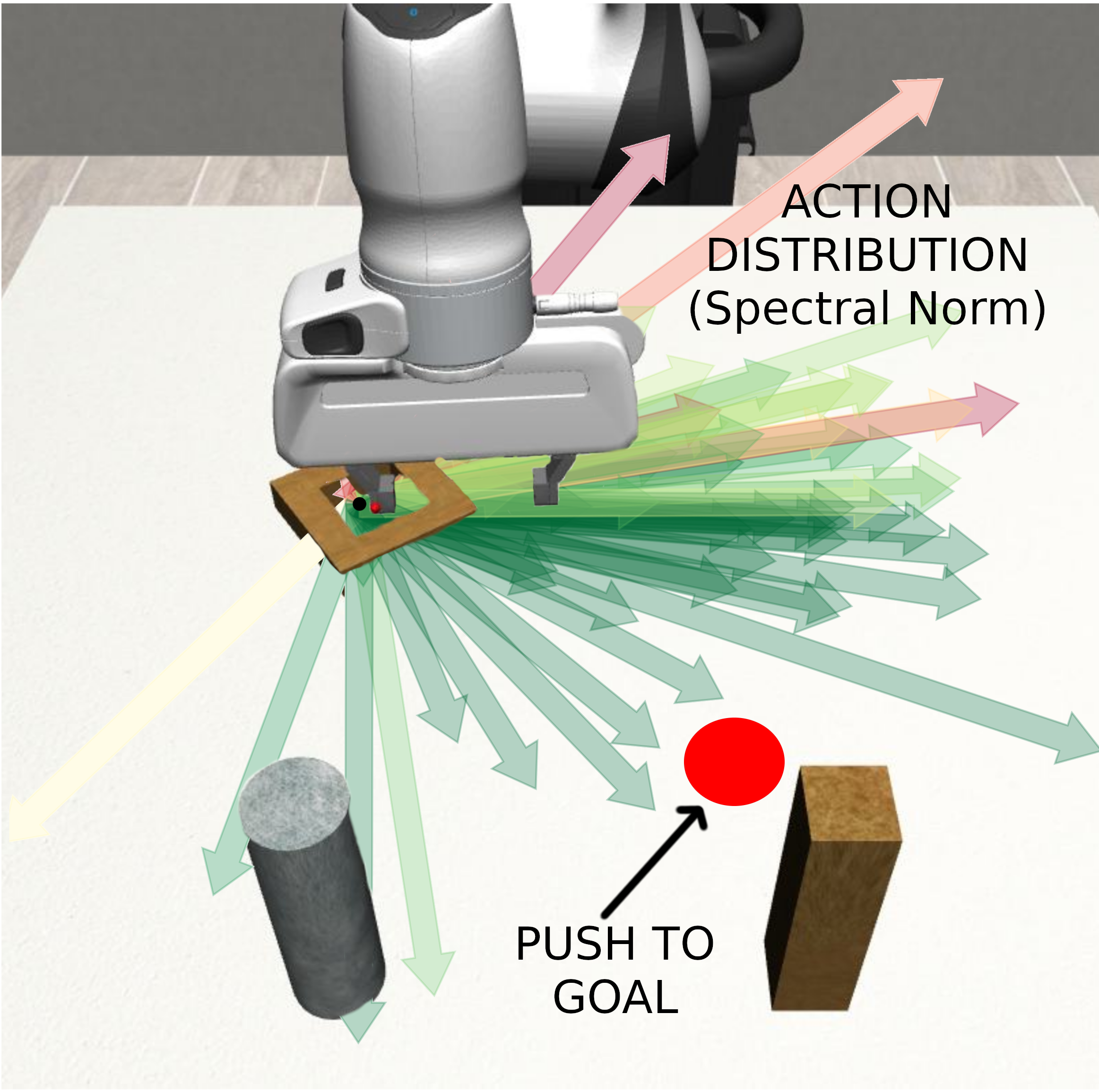}
        \caption{}
        \label{fig:img2}
    \end{subfigure}\hfill%
    \begin{subfigure}[b]{0.29\textwidth}
        \centering
        \includegraphics[width=\textwidth]{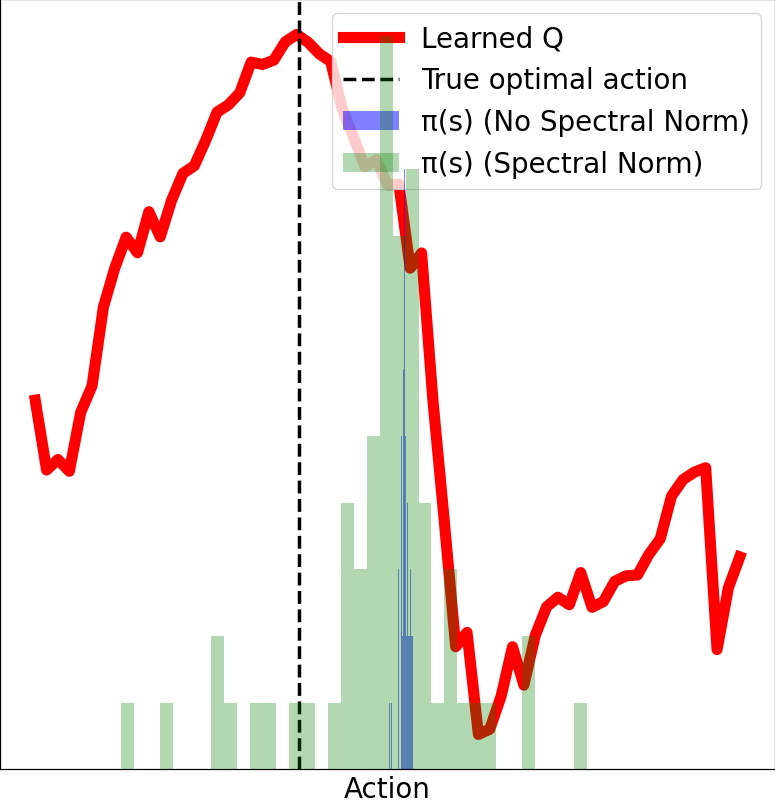}
        \caption{}
        \label{fig:img3}
    \end{subfigure}
    
    \caption{\footnotesize{\textbf{Visualization of the effect of spectral norm}. (a) On a planar pushing task, using IDQL naively results in an excessively narrow marginal action distribution, leading to poor performance. (b) When a spectral norm penalty is added to the behavior policy loss (Equation  \eqref{eq:spectral_norm_loss}), the action distribution is significantly widened. (c) Marginal action distributions projected onto a 1D axis are plotted alongside the learned Q function, which we empirically find to be close to the true Q function in a neighborhood of the data. Narrow action distributions often fail to encompass the optimal action (blue distribution).}}
    \label{fig:stitching_vis}
    \vspace{-2em}
\end{figure*}

How can we train policies in the absence of an explicitly provided reward function? Drawing inspiration from prior work~\cite{reddy2019sqilimitationlearningreinforcement}, we can label all $(o, a)$ transitions in the expert dataset with a reward $r = +1$, while labeling all transitions in the non-expert data $\mathcal{D}_{\text{NE}}$ with reward $r = 0$. We can then use these pseudolabeled datasets to learn policies via a typical offline RL procedure, as described in Section~\ref{sec:background}. The resulting updates become:

\vspace{-15pt}
\begingroup\small
\setlength{\abovedisplayskip}{6pt}
\setlength{\belowdisplayskip}{6pt}
\begin{align} \label{eq:rise_objective}
    \mathcal{L}_V(\psi) &= \mathbb{E}_{(s, a) \sim (\mathcal{D}_{\text{E}} \cup \mathcal{D}_{\text{NE}})} \left[ L^{\tau}_2 (Q_\phi (s, a) - V_{\psi}(s)) \right] \\
    \mathcal{L}_Q(\phi) &= \mathbb{E}_{(s, a) \sim \mathcal{D}_{\text{E}}} \left[ \left( 1 + \gamma V_{\psi}(s') - Q_{\phi}(s, a)\right)^2 \right] \nonumber \\
    &\quad + \mathbb{E}_{(s, a) \sim \mathcal{D}_{\text{NE}}} \left[ \left(\gamma V_{\psi}(s') - Q_{\phi}(s, a)\right)^2 \right] \\
    \pi_B(a|s) &= \text{argmax}_{\pi} \mathbb{E}_{{s, a} \sim (\mathcal{D}_{\text{E}} \cup \mathcal{D}_{\text{NE}})}\left[\log \pi(a|s) \right] \\
    \pi^*(a|s) &= \underset{a \in \{a_1,\dots, a_K\} \sim \pi_B(a|s)}{\text{argmax}} Q_\phi(s, a).
\end{align}
\endgroup

Intuitively, this incentivizes the policy towards state-action transitions in the expert dataset $\mathcal{D}_{\text{E}}$ while using state-action transitions from the non-expert dataset $\mathcal{D}_{\text{NE}}$, to provide a path that returns to the expert state distribution with no additional cost. Since the update in Equation~\ref{eq:rise_objective} performs dynamic programming, it can in principle perform data-efficient ``stitching" of paths from the non-expert data to recover to expert states. While related in spirit to prior work ~\cite{reddy2019sqilimitationlearningreinforcement}, \gOOD is using $0/1$ rewards for continuous action-space offline RL, as opposed to the discrete online RL setting. Naively applying this procedure, however, is insufficient in most robotics problem without an impractically high degree of data coverage, as we show empirically (Fig~\ref{fig:stitching_vis}). Next, we highlight how to practically improve data ``stitchability", allowing policy robustification even in sparse data-coverage setting.

\subsection{Improving Stitchability in Offline Recovery RL} \label{sec:smoothness}

While the methodology in Section \ref{subsec:learning_from_non_expert} should \emph{in principle} stitch behaviors between non-expert and expert data, or stitch within the non-expert data, we find this is not empirically true across several high-dimensional robotic manipulation problems (Table~\ref{tab:sim_results}). Despite having seemingly high-coverage non-expert data, the likelihood of state-overlap in a continuous space tends to $0$, making stitching across exactly overlapping states unlikely. This prevents offline RL from determining paths for recovering to expert states from non-expert ones, even when such paths do exist. While challenging to solve in the most general case, we base our practical improvements on a set of empirical findings in a robotic manipulation setting. 

Empirically, we observe that Q-value functions learned with the expectile regression objective~\cite{idql} tend to be accurate and interpolate well within a neighborhood of the training data, showing reasonable stitching behavior. This is visualized by the solid red line in Fig~\ref{fig:stitching_vis} (c) -- we can see that despite the state-action coverage being incomplete, the landscape of the Q-function in a neighborhood of the training data is accurate -- suggesting that the optimum of the Q-function provides actions that are better than behavior data. However, as shown in prior work, for methods like IDQL, the challenge comes from the \emph{policy extraction} step~\cite{park2024valuelearningreallymain}. While learned Q functions can interpolate in a neighborhood, the marginal action distribution $\pi_B(\cdot | s)$ captured by the behavior policy tends to be overly conservative. The learned distribution overfits to the training set, producing a ``narrow" action distribution that fails to encompass optimal actions (see blue policy distribution in Fig. \ref{fig:stitching_vis}(c)). This prevents trajectories from ``stitching" together even when they might appear to be close, since sampling-based policy extraction is unable to find the optimal action suggested by the ``stitched" value function. 

Given this empirical finding, if we assume the learned Q-function is accurate within a neighborhood of actions in the training distribution, we can achieve better performance by explicitly ``widening" the marginal base policy distribution $\pi_B$. We formalize this notion with the following assumption:

\begin{assumption}
Let $\mathfrak{N}_{d}(a | s) \coloneqq \{ a' \vert  d(a, a' | s) < T \} $ denote the neighborhood of an action~$a$ at state~$s$, i.e., the actions within $T$~distance under distance metric~$d$. Define $J_{\mathcal{D}}(\pi) \coloneqq \mathbb{E}_{a_t \sim \pi(s_t)} \left[ \sum_{t = 0}^H \gamma^t r(s_t, a_t) | s_0 \sim \mathcal{D}] \right]$. Let $\hat \pi(s)~=~\underset{a \sim \mathfrak{N}(a_0 | s), a_0 \sim \pi_B(s)}{\text{argmax}} Q_\phi(s, a)$. Then, for any $\delta > 0$, there exists $\mathfrak{N}_d$ such that $\vert J_{\mathcal{D}}(\pi) - J_{\mathcal{D}}(\pi_{opt}) \vert < \delta$, where $\pi_{opt}$ is the optimal policy.
\end{assumption}

We find that a natural way to choose such a neighborhood to widen the action distribution of $\pi_B$, is to \emph{alias} action distributions between nearby states. In doing so, there is a natural notion of ``fuzziness" that is introduced between nearby states, preventing the overly conservative policy behavior mentioned above. We focus on two techniques here: 

\textbf{Enforcing Policy Lipschitz Continuity:} One way to implicitly induce aliasing between action distributions at nearby states is to enforce Lipschitz continuity on the policy $\pi_B$. This ensures that action distributions at nearby states are similar, avoiding overly conservative action distributions. While there are several ways to enforce Lipschitz continuity, we opt for regularizing the policy with a spectral norm penalty~\cite{ke2024ccilcontinuitybaseddataaugmentation, yoshida17spectral}

\vspace{-15pt}
\begingroup\small
\setlength{\abovedisplayskip}{10pt}
\setlength{\belowdisplayskip}{10pt}
\begin{align} 
\label{eq:spectral_norm_loss}
    \max_\theta \, \mathbb{E}_{(s, a) \sim (\mathcal{D}_{\text{E}} \cup \mathcal{D}_{\text{NE}})}\!\left[ \log \pi_\theta(a|s) \right] + \lambda \!\sum_{W \in \theta} \!\|\sigma_{\text{max}}(W)\|^2.
\end{align}
\vspace{-15pt}
\endgroup

\noindent Spectral normalization has been shown to bound the Lipschitz constant of a learned model \cite{O_Connell_2022}. This objective is simple to optimize using gradient-based supervised learning procedures. 

% In the concrete instantiation of this work, we replace the maximum likelihood objective with the standard DDPM denoising objective to train diffusion models ~\cite{hodiffusion, chi23diffusion}. 

\textbf{Distance-Based Data Augmentation:} An explicit method of widening the distribution of $\pi_B$ is to augment $\mathcal{D}_U$ with additional transitions in the neighborhood. For a given $(s, a)~\in~\mathcal{D}_U$, we choose $\mathfrak{N}_d(a | s)$ to be actions from states close to $s$, as specified by the distance metric $d$. For every pair of transitions $(s, a), (s', a') \in \mathcal{D}_{\text{E}} \cup \mathcal{D}_{\text{NE}}$, we construct an augmented dataset $\mathcal{D}_{\text{aug}}$ by adding $(s, a')$ to $\mathcal{D}_{\text{aug}}$ if $d(s, s')~<~T$ for some distance metric $d$ and threshold $T$, 

\vspace{-10pt}
\begingroup\small
\begin{equation}
    \mathcal{D}_{\text{aug}} = \{ (s, a') | (s, a), (s', a') \in \mathcal{D}_{\text{E}} \cup \mathcal{D}_{\text{NE}} \text{ if } d(s, s') < T\}.
\end{equation}
\endgroup
\vspace{-15pt}

We then train $\pi_B$ via supervised learning on the entire augmented dataset $\mathcal{D}_{\text{E}} \cup \mathcal{D}_{\text{aug}}$, to learn a broader marginal policy distribution. While the choice of distance metric can vary, we find that using an Euclidean distance in the feature space of a large pretrained vision model, DINOv2 \cite{oquab2024dinov2learningrobustvisual}, which has been shown to measure meaningful semantic differences between images, is effective. The version of \gOOD in our experimental evaluation has both spectral norm penalty and distance-based data augmentation included. As we show experimentally, these additions make a significant difference in the ability of \gOOD to use non-expert data for robust policy learning. In summary, \gOOD provides a simple way to augment imitation learning policies with a critic learned via expectile regression to effectively make use of non-expert data for recovery and broad generalization, even in the low data-coverage regime. 

% \footnote{This particular choice of distance metric was effective in our experiments, but can likely be improved in future work.}
% ====================== 05 Experimental Setup ======================
% \input{figures/01_long_subfigs_of_envs}

\section{Experimental Setup}

\begin{figure}[!t]
    \centering
    \includegraphics[width=\columnwidth]{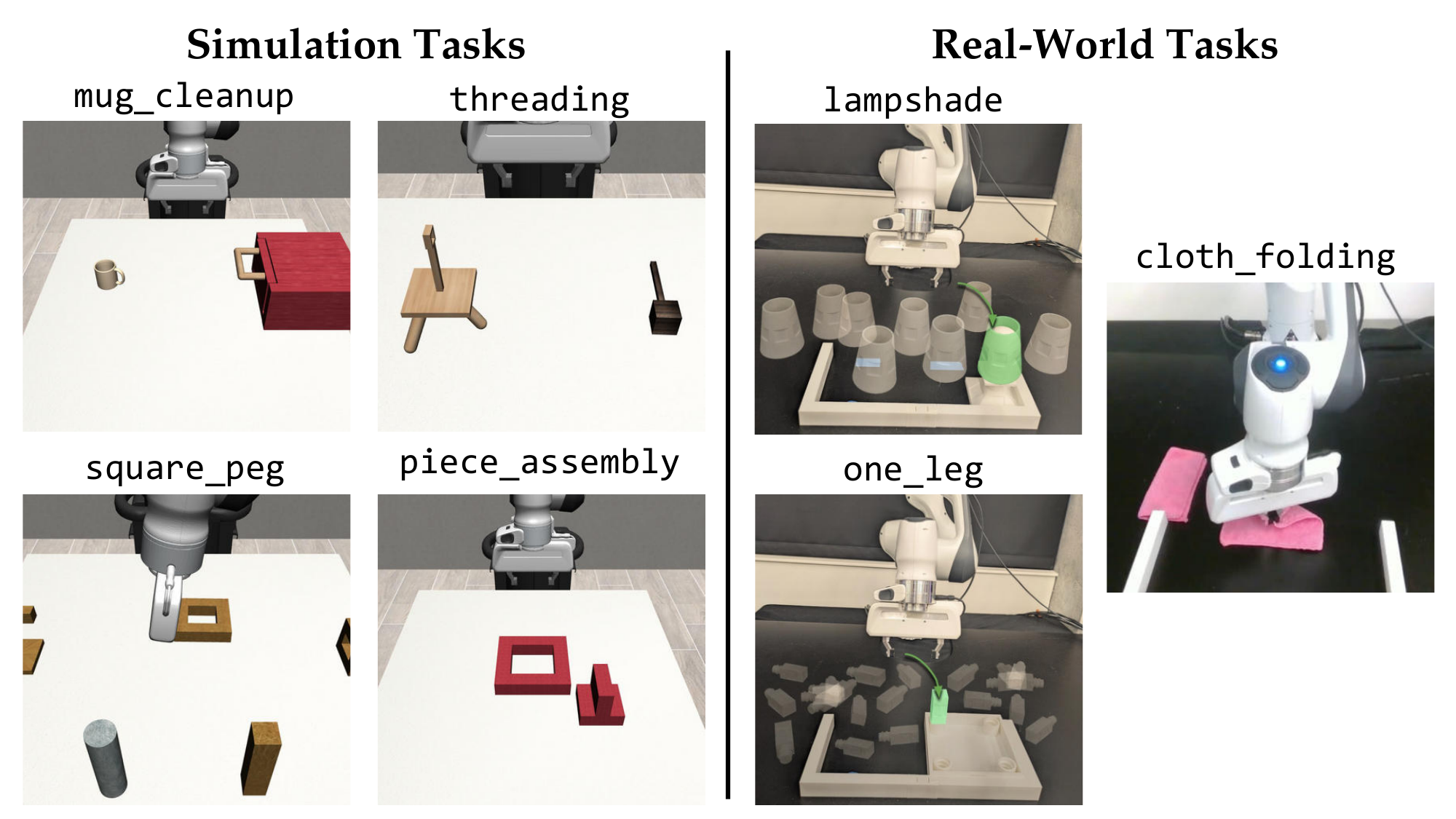}
    \caption{\footnotesize{Depiction of tasks in sim and the real world.}}
    \label{fig:envfigs}
\end{figure}

\textbf{Evaluation Tasks:} We investigate the \gOOD approach on manipulation tasks in simulation from the Robomimic benchmark~\cite{robomimic, mandlekar2023mimicgendatagenerationscalable}, and real-world robot tasks from the Furniture Bench \cite{heo2023furniturebench} benchmark (as shown in Fig \ref{fig:envfigs}). We choose tasks that cover a range of characteristics including $\mathbb{SE}(2)$ object rearrangement (\lamp), $\mathbb{SE}(3)$ object manipulation (\squarenut, \twopiece), fine-precision (\threading, \cloth), and long-horizon behavior (\mug, \oneleg).  This work is evaluated on the Franka Panda robot both in simulation and the real world. In all evaluations, the policies and value functions receive camera images (from both wrist and third person cameras), as well as proprioceptive joint state from the robot. We refer the reader to the Appendix for task/implementation details. 

\textbf{Evaluation Settings:} We consider three evaluation ``settings" - (1) learning to recover from unstructured play data, (2) leveraging suboptimal failure data to improve success rates, and (3) iteratively improving a policy by finetuning on its own evaluation rollouts.  For each task, we collect a set of expert demonstrations completing the task from a range of initial object configurations, and a set of non-expert demonstrations, which have different qualities for each setting, as we outline below. See Fig~\ref{fig:2piece_vis} for a visualization of the training data.

(1) \emph{Learning to recover from unstructured play data:} This involves scenarios where a set of expert data that completes the task is augmented with a larger set of human collected, unstructured, undirected ``play" data. This data demonstrates how to move the object around in the environment, thereby enabling recovery from unfamiliar starting conditions. For instance, in Fig \ref{fig:recoveryfig}, while the expert (shown in green) can only succeed from a narrow region, the undirected data (shown in yellow) can enable recovery back to this region to solve the task reliably across the state space. This type of data is typically very cheap to collect, as it does not require the precision to complete a task.

(2) \emph{Leveraging suboptimal failure data:} This involves scenarios where a set of expert data that completes the task is augmented with a larger set of human collected suboptimal or failed demonstrations, which often occurs naturally during data collection. While the failed demonstrations are not suitable for direct imitation, they can still demonstrate  useful subcomponents  of the task. When these are stitched together with expert behavior, this leads to robust, higher-coverage policies, without wasting the entirety of the suboptimal data. 

(3) \emph{Iterative Policy Improvement:} We also demonstrate that useful non-expert data can be collected from policy rollouts, not just human demonstrations. Like in setting (1), given an initial expert dataset $\mathcal{D}_E$ and non-expert dataset $\mathcal{D}_{NE}$, we train a policy $\pi^*$ as given in Equation \ref{eq:rise_objective}. We then evaluate the policy $\pi^*$, and add successful rollouts to $\mathcal{D}_E$ and failed rollouts to $\mathcal{D}_{NE}$, then re-train.

We consider several imitation and offline RL baselines - (1) \emph{Behavior cloning:} This is the standard imitation learning paradigm, with a diffusion policy~\cite{chi23diffusion} trained on only the expert dataset $\mathcal{D}_{\text{E}}$, (2) \emph{Behavior cloning unified:} This is similar to behavior cloning, but on the union of expert and non-expert data $\mathcal{D}_{\text{E}} \cup \mathcal{D}_{\text{NE}}$,  (3) \emph{ILID:} This is an implementation of the data filtering algorithm in~\cite{ilid}, where a classifier is used to classify expert vs non-expert states and subtrajectories that have overlap with expert data are selectively added to the training dataset for imitation learning, (4) \emph{SQIL}: an online RL method that originally proposed 0/1 rewards, implemented as offline SAC \cite{reddy2019sqilimitationlearningreinforcement}, (5) \emph{CQL}: a common offline RL method that enforces conservatism on the Q function \cite{cql}, and (6) \emph{IDQL}: the method RISE builds off of, without any data augmentation or Lipschitz penalty \cite{idql}. We modify the original IDQL implementation to use the $0/1$ rewards proposed in \gOOD.   

\section{Results}
\label{sec:experimental_results}

\begin{table*}[!t]
\vspace{-5pt}
\centering
\footnotesize % Reduce font size
\setlength{\tabcolsep}{3pt} % Reduce column separation significantly
\begin{tabular}{>{\raggedright\arraybackslash}c | l | c | c | c | c | c | c | c}
\hline
\textbf{Data Type} & \textbf{Sim Task Variant} & \textbf{BC} & \textbf{BCU} & \textbf{ILID} & \textbf{SQIL} & \textbf{CQL} & \textbf{IDQL} & \textbf{\gOOD}  \\
\hline
\multirow{5}{*}{{\footnotesize Coverage}} & \squarenut  & $18.7 \pm 2.4$ & $0.0 \pm 0.0$ & $35.3 \pm 3.5$ & $0.0 \pm 0.0$ & $12.4 \pm 3.2$ & $19.6 \pm 4.3$ & $\mathbf{50.7 \pm 5.8}$ \\
& \squarehook & $18.0 \pm 3.5$ & $0.0 \pm 0.0$ & $34.6 \pm 2.4$ & $0.0 \pm 0.0$ & $10.5 \pm 3.9$ & $17.8 \pm 5.8$ & $\mathbf{47.9 \pm 1.2}$ \\
& \twopiece  & $14.7 \pm 2.9$ & $2.0 \pm 1.2$ & $43.3 \pm 2.4$ & $3.3 \pm 2.4$ & $8.2 \pm 1.5$ & $16.3 \pm 2.0$ & $\mathbf{70.7 \pm 8.8}$ \\
& \twopiece (tip) & $0.0 \pm 0.0$ & $0.0 \pm 0.0$ & $9.3 \pm 1.3$ & $0.0 \pm 0.0$ & $0.0 \pm 0.0$ & $8.0 \pm 2.3$ & $\mathbf{51.3 \pm 9.3}$ \\
& \threading  & $17.3 \pm 2.6$ & $0.0 \pm 0.0$ & $20.3 \pm 1.9$ & $0.0 \pm 0.0$ & $0.0 \pm 0.0$ & $9.8 \pm 3.9$ & $\mathbf{22.7 \pm 1.4}$ \\
\hline
\multirow{3}{*}{{\footnotesize Suboptimal}}& \mug & $31.3 \pm 3.5$ & $32.7 \pm 1.8$ & $24.7 \pm 4.1$ & $6.0 \pm 1.3$ & $22.3 \pm 2.0$ & $36.7 \pm 3.2$ & $\mathbf{40.7 \pm 5.3}$ \\
& \twopiece (tip) & $20.0 \pm 3.1$ & $23.3 \pm 5.7$ & $22.7 \pm 2.4$ & $0.0 \pm 0.0$ & $16.7 \pm 2.1$ & $\mathbf{35.7 \pm 4.5}$ & $\mathbf{36.0 \pm 6.1}$ \\
& \squarenut  & $8.0 \pm 2.3$ & $34.0 \pm 2.0$ & $32.0 \pm 2.3$ & $8.3 \pm 1.7$ & $25.3 \pm 2.2$ & $41.3 \pm 8.2$ & $\mathbf{56 \pm 2.3}$ \\ 
\hline
\hline
\textbf{Data Type} & \textbf{Real Task Variant} & \textbf{BC} & \textbf{BCU} & \textbf{ILID} & \textbf{SQIL} & \textbf{CQL} & \textbf{IDQL} & \textbf{\gOOD} \\
\hline
\multirow{2}{*}{\footnotesize {Coverage}}
 & \lamp & $17.5$ & $45.0$ & $57.5$ & $0.0$ & $0.0$ & $10.0$ & $\mathbf{82.5}$ \\
 & \cloth & $0.0$ & $8.0$ & $12.0$ & $0.0$ & $0.0$ & $16.0$ & $\mathbf{24.0}$ \\
\hline
\multirow{1}{*}{\footnotesize Suboptimal}
& \oneleg  & $25.0$ & $0.0$ & $30.0$ & $0.0$ & $0.0$ & $0.0$ & $\mathbf{50.0}$ \\
\hline
\end{tabular}
    \caption{\footnotesize{\textbf{Sim \& real tasks across benchmarks:}
    Success percentage for an array of tasks with different types of human collected non-expert data. \squarenut and \squarehook share the same non-expert data.}. Coverage refers to experiments utilizing high coverage play data (setting (1)), while suboptimal refers to experiments utilizing suboptimal failure data (setting (2)).}
     \label{tab:sim_results}
\vspace{-15pt}
\end{table*}

\paragraph{\textbf{\gOOD solves tasks from a broad range of initial configurations using high-coverage play data:}} 

\begin{figure}[!h]
    \centering
    \begin{subfigure}[b]{0.48\columnwidth}
        \centering
        \includegraphics[width=\textwidth]{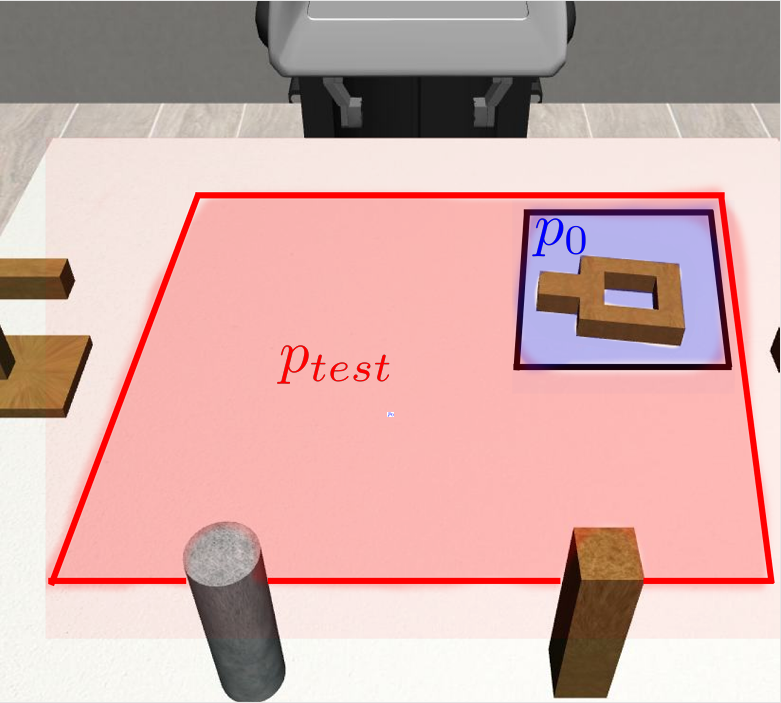}
        \label{fig:p0viz}
    \end{subfigure}\hfill%
    \begin{subfigure}[b]{0.48\columnwidth}
        \centering
        \includegraphics[width=\textwidth]{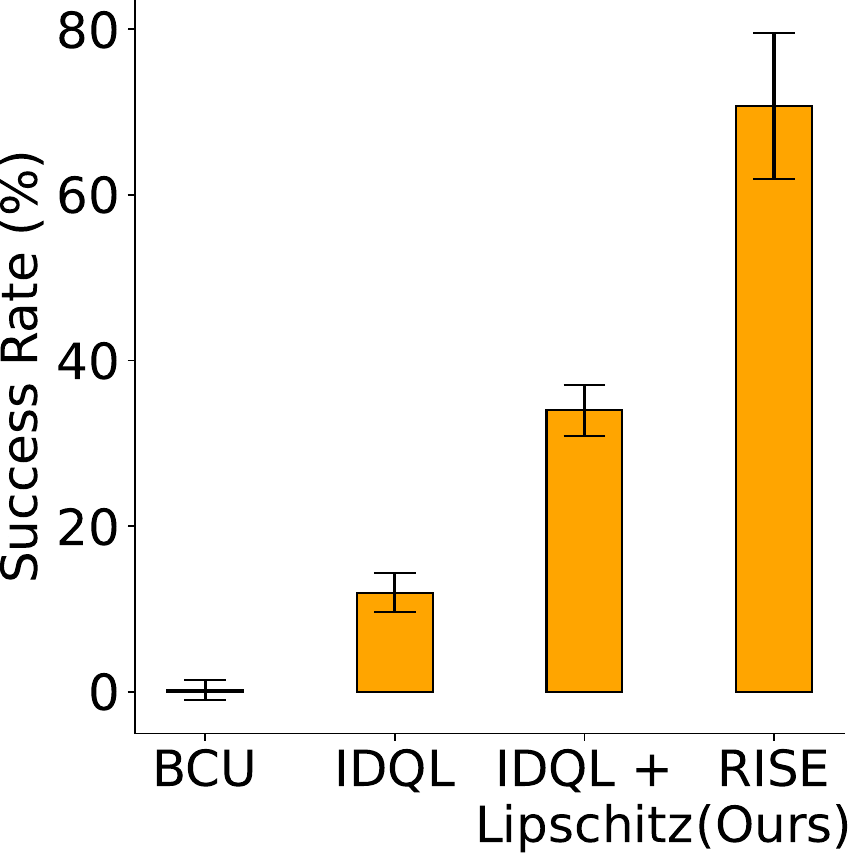}
        \label{fig:lipschitz_ablation}
    \end{subfigure}
    \vspace{-5pt}
    \caption{ \footnotesize{\textbf{Generalization and Ablation} (a) All experts are demonstrated from a narrow initial distribution $p_0$. We test in a larger region $p_{test}$. Our method is able to generalize to $p_{test}$ only using cheap play data. (b) Ablation of applying spectral norm regularization and data augmentation to standard IDQL on \twopiece with high coverage. Standard offline RL (IDQL) does not stitch well, but adding our lipschitz constraint and data augmentation greatly improves performance.}}
    \label{fig:combined_6_7}
    \vspace{-5pt}
\end{figure}

With the addition of low collection cost play data (as shown in Fig \ref{fig:recoveryfig}) to just expert data, our results indicate that \gOOD is able to achieve strong performance on a much wider distribution of initial configurations than an expert policy naively trained with imitation learning (Figure \ref{fig:combined_6_7}a). This can be seen from the improvement of \gOOD over BC in both simulation and real (See Coverage section in Table \ref{tab:sim_results}). Crucially, we do not have to demonstrate expert behavior from this wider distribution, but simply collect enough coverage data which can be \emph{stitched} with the expert data to enable recovery to the expert manifold. The BCU results suggest that simply imitating the high-coverage play data is insufficient, and this needs to be used in a targeted way. While ILID~\cite{ilid}, along with the other offline RL baselines (SQIL, CQL, IDQL), can utilize suboptimal data to some extent, they are poor at stitching trajectories together, making them far less effective than \gOOD across tasks. This gap is particularly pronounced for the \twopiece with tipping task, which requires combining multiple behaviors together (first rotating the object, then recovering). Notably, these results hold across both simulation and real world tasks. Moreover, since the non-expert data is simply used to recover back to the expert manifold, the same non-expert data can be useful across multiple downstream tasks. In Table~\ref{tab:sim_results}, \gOOD achieves good perfomance on the \squarehook and \squarenut tasks, which share the same non-expert data. This shows that the same data can be stitched to two different experts, offering a scalable way of improving policy robustness.

\paragraph{\textbf{\gOOD is able to use suboptimal or partial data to improve policy performance:}}

Our results show that \gOOD is able to utilize suboptimal or partial trajectory data to improve evaluation performance of the resulting policy (Table~\ref{tab:sim_results} under the Suboptimal data type section). While simply imitating a mixture of suboptimal data and optimal data leads to a considerable drop off in imitation learning methods (BCU), \gOOD is able to filter out the suboptimal data and do significantly better. Moreover, we see that \gOOD is actually able to outperform the standard BC baseline, which is simply imitating the expert data (while discarding suboptimal data). This suggests that \gOOD is not only filtering the data to ignore poor demonstrations, but also stitching suboptimal with optimal data to see additional benefit. As before, ILID~\cite{ilid} can show some benefit, but generally does not make maximal use of the suboptimal data because of its inability to stitch together data. 

\paragraph{\textbf{\gOOD is able to leverage data collected from policy evaluations:}} 

Policy evaluations are run frequently in the real world, and provide a rich source of additional data. While not typically used in the learning pipeline, we show that iteratively re-integrating this evaluation data into policy learning can help. Table \ref{tab:iterative_improvement} shows that \gOOD is able to leverage data collected from policy evaluation to improve policy performance without any additional human demonstrations. Given an initial policy that performs relatively poorly at the task, we are able to use the data from the rollouts from that very same policy to improve the policy by categorizing them as either successful trajectories or failures. With each subsequent round of data collection and re-training, we see that the policy performance increases. This demonstrates the versatility of \gOOD in utilizing all forms of non-expert data. 

\begin{table}[!h]
    \centering
    \begin{tabular}{c|c|c|c}
        \hline
        Task & Initial Performance & Iteration 1 & Iteration 2 \\
        \hline
        \twopiece & 26.3 & 42.7 & 49.0 \\
        \lamp & 20.0 & 55.0 & 60.0 \\
        \hline
    \end{tabular}
    \caption{\footnotesize{Results for iterative policy improvement using data collected autonomously from policy rollouts. Given a poor initial policy, additional data is collected from its rollouts to finetune the policy. This process is repeated over multiple iterations.}}
    \label{tab:iterative_improvement}
\vspace{-5pt}
\end{table}

\paragraph{\textbf{Ablations and Analysis}} 

\emph{\underline{Impact of Lipschitz continuity and Data Augmentation:}} To understand the impact of imposing Lipschitz continuity on \gOOD and data augmentation, we also perform a targeted ablation on the \twopiece task in simulation. From Fig~\ref{fig:combined_6_7}b, we can see that offline RL for recovery (without any smoothness additions) performs better than naively doing BC, but can be improved by imposing of Lipschitz continuity through the spectral norm. Fig.~\ref{fig:combined_6_7}b further shows performance gains by adding the distance-based data augmentation. 

\begin{figure}[!h]
    \centering
    % First row
    \begin{subfigure}[b]{0.2\textwidth}
        \centering
        \adjustbox{width=\textwidth, height=3.5cm, keepaspectratio=false}{%
            \includegraphics{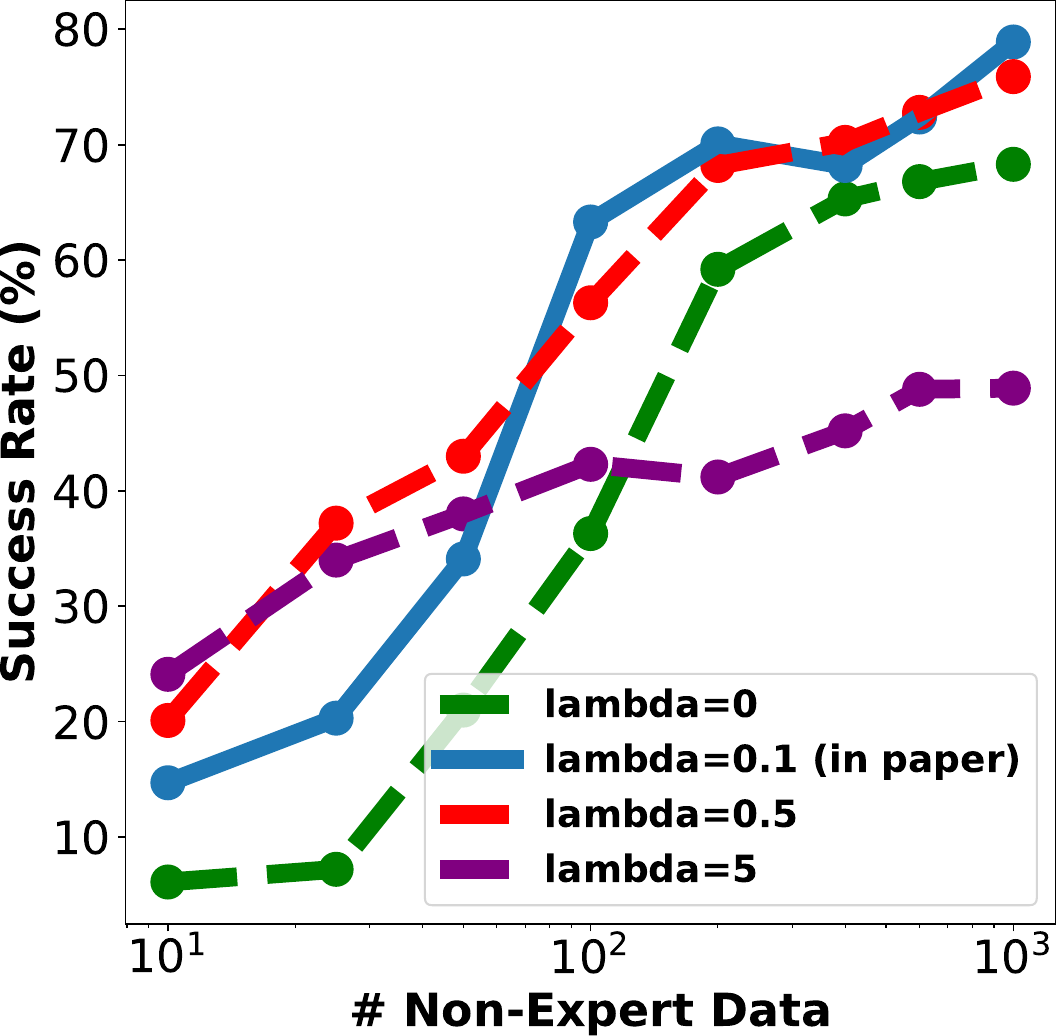}%
        }
        \caption{}
        \label{fig:img1_ablations}
    \end{subfigure}%
    \begin{subfigure}[b]{0.2\textwidth}
        \centering
        \adjustbox{width=\textwidth, height=3.5cm, keepaspectratio=false}{%
            \includegraphics{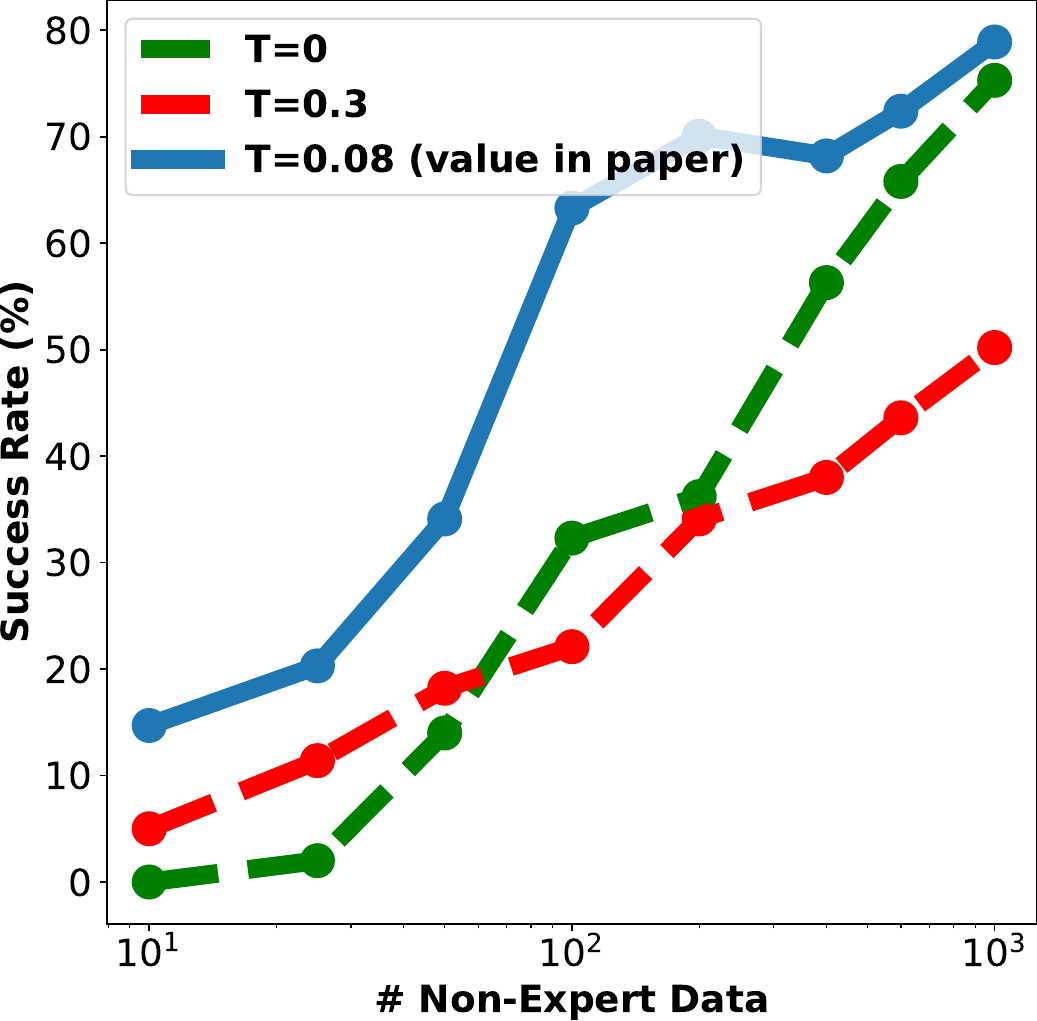}%
        }
        \caption{}
        \label{fig:img2_ablations}
    \end{subfigure}%
    \caption{\footnotesize {Ablations on relation between data quantity and (a) $\lambda$ and (b) $T$ hyperparameters for the \twopiece task. Moderate spectral normalization penalty and data augmentation, which expands the policy's action distribution, is critical, particularly when data is scarce.}}
    \label{fig:ablations}
    \vspace{-10pt}
\end{figure}

\emph{\underline{Impact of smoothing hyperparameters:}} We examine the effect of various parameters of $\lambda$, the strength of the spectral norm regularization, $T$, the distance threshold governing the degree of data augmentation, and $| \mathcal{D}_{NE} |$, the amount of non-expert data. In, Fig~\ref{fig:ablations} (a) and (b), we see the sensitivity of RISE with respect to $\lambda$ and $T$, respectively on the \twopiece task, and its relation to the amount of data. We see that a moderate amount of spectral normalization and data augmentation greatly increases policy success, and as expected, this improvement is greatest when data is limited. The performance is somewhat sensitive to hyperparameter values, but a large range of values is beneficial.

\section{Conclusion and Limitations}

\gOOD provides a new way to use ideas from offline RL to improve the robustness of imitation learning, but without requiring the challenging reward labeling procedure involved in most offline RL methods. Informally, key insight here is to ``fuzz'' the dataset in places where precision is not required using a notion of Lipschitz continuity. With this, however, comes a caveat: You must know which parts of the dataset needs to be precise and which parts can sacrifice precision for stitch-ability. In some settings, it is clear, while in others this may require more careful tuning. We also find that there are scenarios where the suboptimal and optimal data do not overlap, despite the smoothing offered by Lipschitz continuity and data augmentation. A clear understanding of what data sources will yield benefits would be valuable in future studies. 
\section{Acknowledgements}
The authors would like to acknowledge members of the Robot Learning Lab and the Washington Embodied Intelligence and Robotics Development Lab for helpful and informative discussions throughout the process of this research. The authors would also like to thank Emma Romig at the University of Washington for their help in setting up the robotic hardware and teleoperation interfaces for this project. This research was supported by funding from Toyota Research Institute, under the University 2.0 research program. 

% Use BibTeX with IEEE-compatible style
\bibliographystyle{ieeetr}
\bibliography{references} 

% Appendix comes after references
\newpage
\appendix
\section{Ablations} \label{appendix:ablations}

\section{Implementation Details} \label{appendix:impl_details}

\begin{figure}[!h]
    \centering
    \begin{subfigure}[b]{0.22\textwidth}
        \centering
        \includegraphics[width=\textwidth]{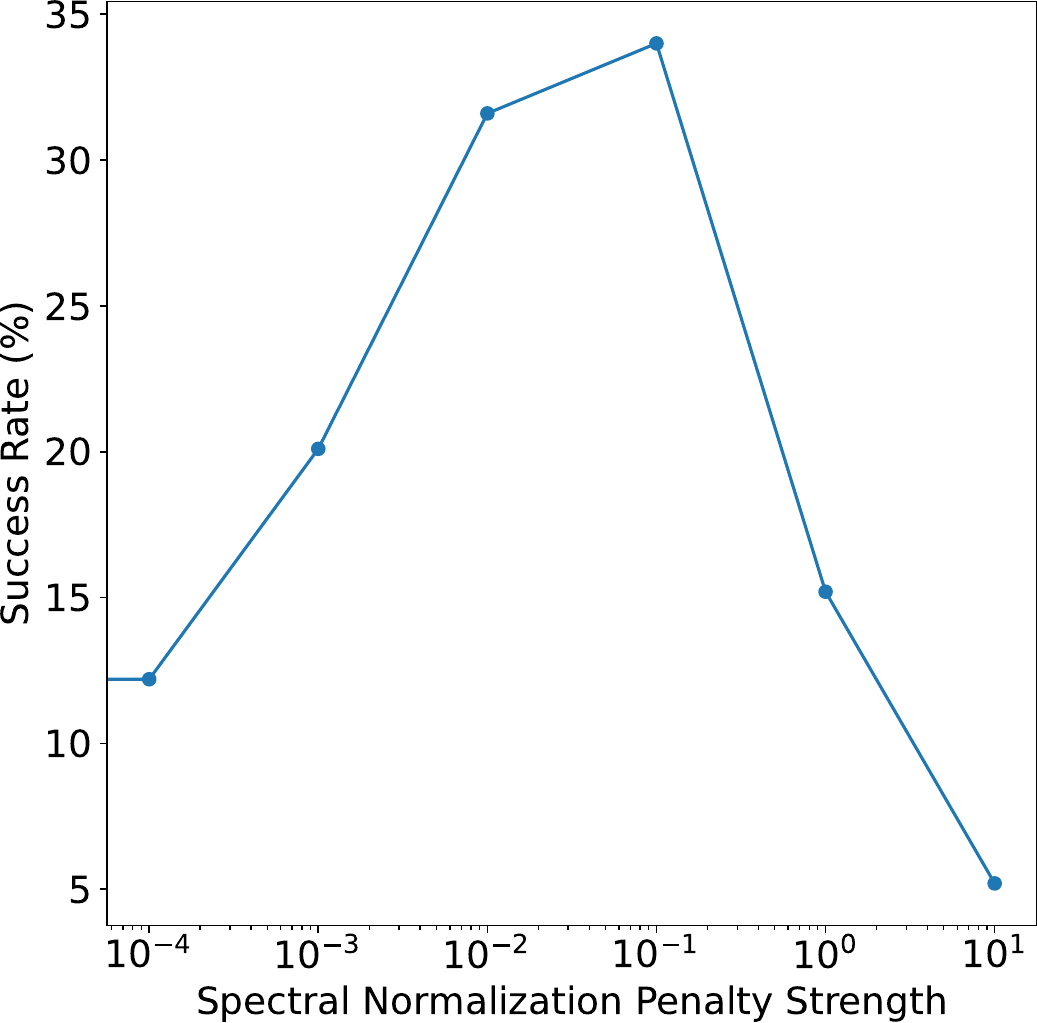}
        \caption{}
    \end{subfigure}
    \begin{subfigure}[b]{0.22\textwidth}
        \centering
        \includegraphics[width=\textwidth]{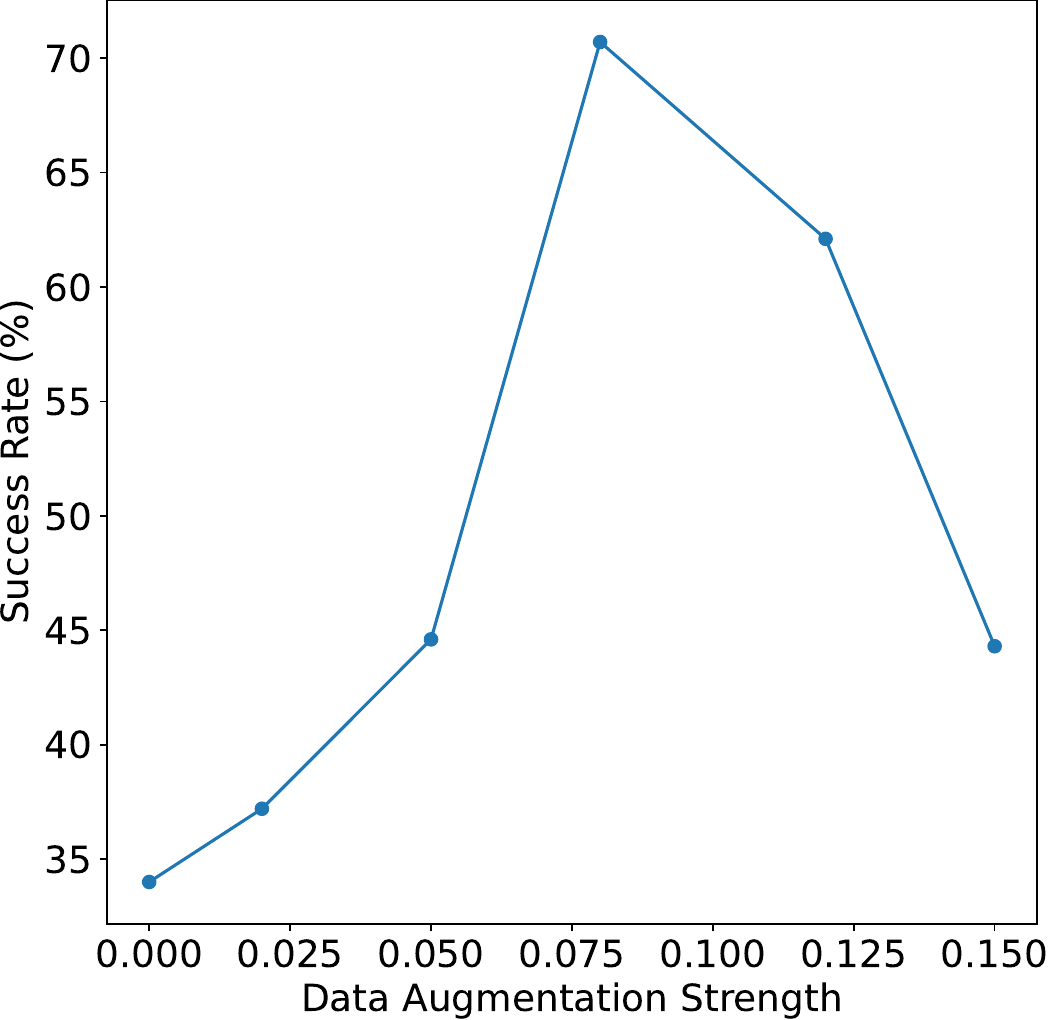}
        \caption{}
    \end{subfigure} \\
    \caption{\footnotesize{\textbf{Ablation on piece assembly task} Effect of various (a) spectral normalization penalty strength parameters $\lambda$ and (b) data augmentation threshold parameters $T$ on the piece assembly task. Spectral normalization is applied assuming no data augmentation, while data augmentation ablations are done using the optimal level of spectral normalization $\lambda = 0.1$}}
    \label{fig:2piece_ablation}
\end{figure}

\textbf{Additional Ablations} We provide a few more examples of ablations into the hyperparameters $\lambda$ and $T$. First, looking at the \twopiece task, we see that a moderate amount of spectral normalization and data augmentation greatly increases policy success, as seen in Figure \ref{fig:2piece_ablation}. From this ablation, we use the best value of $\lambda$ ($\lambda = 0.1$) for all experiments utilizing high coverage non-expert datasets in simulation. Data augmentation strength, however, has to be tuned per-task.

\begin{figure}[!h]
    \centering
    \begin{subfigure}[b]{0.22\textwidth}
        \centering
        \includegraphics[width=\textwidth]{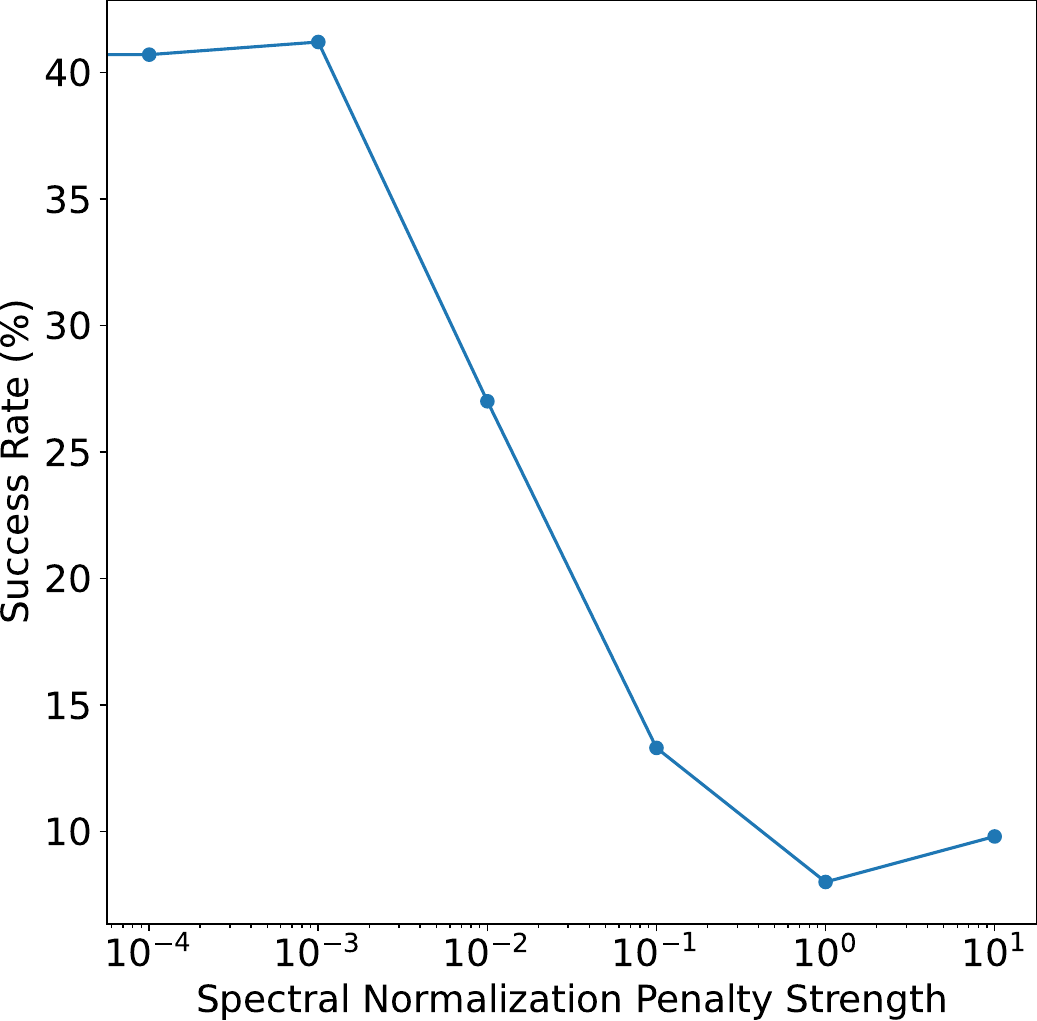}
        \caption{}
    \end{subfigure}
    \begin{subfigure}[b]{0.22\textwidth}
        \centering
        \includegraphics[width=\textwidth]{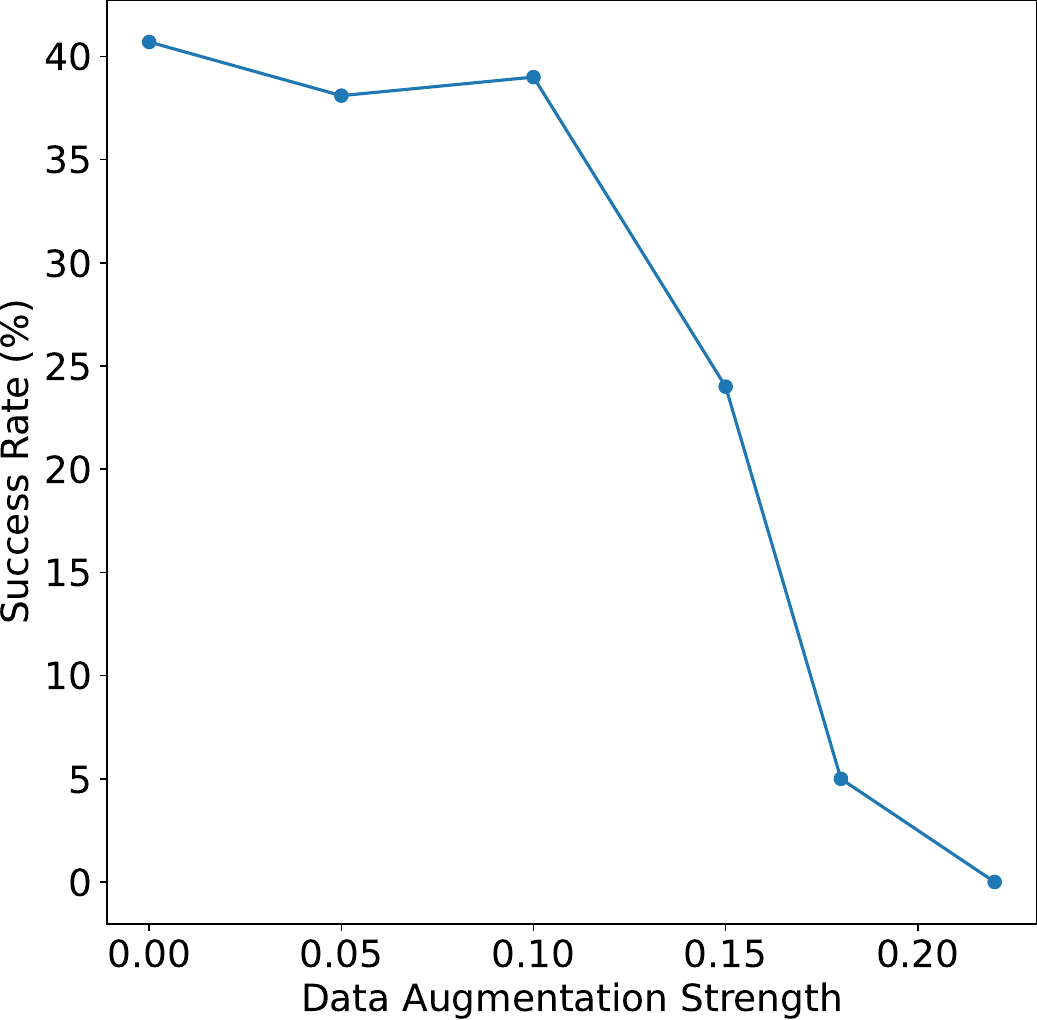}
        \caption{}
    \end{subfigure} \\
    \caption{\footnotesize{\textbf{Ablation on mug cleanup task} Effect of various (a) spectral normalization penalty strength parameters $\lambda$ and (b) data augmentation threshold parameters $T$ on the mug cleanup task. Spectral normalization is applied assuming no data augmentation, while data augmentation ablations are done using the optimal level of spectral normalization $\lambda = 0.001$}}
    \label{fig:mug_ablation}
\end{figure}

We use the \mug task as representative of experiments where we use suboptimal and partial demonstrations. For these datasets, we use the best value of $\lambda = 0.001$ demonstrated in Figure \ref{fig:mug_ablation}. 

\textbf{Experimental Setup} All simulation task environments are modified from versions implemented in Robomimic or Mimicgen \cite{robomimic,mandlekar2023mimicgendatagenerationscalable}, which are built on the Mujoco simulator \cite{todorov2012mujoco}. Data is collected with a SpaceMouse device.  Real experiments are conducted on a Franka Panda robot. We collect robot trajectory demonstrations using the Gello leader arm \cite{wu2023gello}. During collection, the Panda follows them using the waypoint-tracking impedence controller provided by \texttt{libfranka} \cite{libfranka} and the \texttt{franky} wrapper \cite{franky}. In both simulation and real, the action space consists of delta cartesian end-effector commands. The observation consists of proprioception from the robot, in the form of the cartesian end effector position, the orientation of the end effector as a quaternion, and the gripper state. The observation also contains two RGB images, one from a fixed exocentric camera, and one wrist mounted camera. In real, the camera images are captured from two Intel Realsense D435 cameras. 

\textbf{Architecture and Training} The behavior policy is parametrized as a diffusion policy with a U-Net denoiser, with the same implementation and parameters as \cite{chi23diffusion}. The value network and Q network are each a 3-layer MLP. We use the ResNet18 architecture \cite{he2015deepresiduallearningimage} for the image encoder, which is trained jointly with the behavior policy and Q function. Like in \cite{chi23diffusion}, we use \textit{action chunking}, where the policy predicts 
 a sequence of actions \cite{zhao23act}. The behavior policy also takes as input an observation history, where the observations from the last two timesteps are stacked together. For all tasks, each network is trained for 5000 epochs.

\textbf{Data Augmentation} For computational efficiency, we approximate the data augmentation procedure described in Section \ref{sec:smoothness} using a k-nearest neighbors algorithm. For each state, we compute the $k = 15$ nearest neighbors under the distance metric $d$. If the distance is less than the threshold $T$, and the two states belong to different trajectories, we accept the the states, swapping their actions. To compute the distance metric, we use the ViT-S model from DinoV2 \cite{oquab2024dinov2learningrobustvisual}, extract the patch tokens, normalize, and then compute euclidean distance. 

\textbf{Hyperparameters} Hyperparameters used are provided in Table \ref{tab:hyperparams}. Data augmentation strength $T$ is tuned per-task. Data quantities for each task is found in Table \ref{tab:datasets}.

\begin{table}[htbp]
    \centering
    \begin{tabular}{c|c}
        \hline
        Parameter & Value \\
        \hline
        Learning Rate (all) & 1e-4 \\
        Batch Size & 64 \\
        Diffusion Timesteps & 100 \\
        Beta Schedule & cosine \\
        Discount Factor $\gamma$ & 0.99 \\
        IQL Expectile $\tau$ & 0.9 \\
        \# Samples from Behavior Policy & 64 \\
        Spectral norm regularization strength $\lambda$ & \makecell{0.1 (coverage experiments) \\ 0.001 (suboptimal experiments) \\ 0.01 (real)}\\
        Action chunking horizon & 16 \\
        Observation history horizon & 2 \\
    \end{tabular}
    \caption{\footnotesize {Table of hyperparameters used.}}
    \label{tab:hyperparams}
\end{table}

\section{Task and Dataset Details} \label{appendix:dataset}

We provide a description of each task and associated datasets. A summary of the data used for each task can be found in Table \ref{tab:datasets}. For the results in Table \ref{tab:sim_results}, sim experiments were evaluated for 100 rollouts over 3 seeds, the \lamp and \oneleg tasks were evaluated for 40 rollouts, and the \cloth task for 25.

\textbf{\squarenut} The task involves picking up a square nut and placing it over a peg. In the expert data, the square nut is initialized in a small region $p_0$ on the upper right of the table. For the ``high-coverage play" data, the nut is randomized across the entire table $p_{test}$, and randomly picked up and moved around in 3D space. The ``suboptimal" data is a subset of Robomimic's ``Square-Worse" dataset for the square task \cite{robomimic}, where an inexperienced human demonstrator attempts to complete the task. For the coverage experiment, we evaluate with the nut initialized to $p_{test}$. For the suboptimal experiment, we evaluate with the nut initialized from $p_0$.

\textbf{\squarehook} The task exists in the same environment as \squarenut, except the goal is to hang the square nut onto a hook. The initial distribution of the nut in the expert data is also identical to \squarenut. We re-use the same high coverage data as the \squarenut task to demonstrate the scalability of this type of data to augment multiple downstream tasks. 

\begin{figure}[!htb]
    \centering
    \setlength{\tabcolsep}{1pt} % Reduce space between columns
    \begin{tabular}{cc} % Use a tabular environment with two columns
        \begin{subfigure}[b]{0.15\textwidth}
            \centering
            \includegraphics[width=\textwidth,height=2.0cm,keepaspectratio=false]{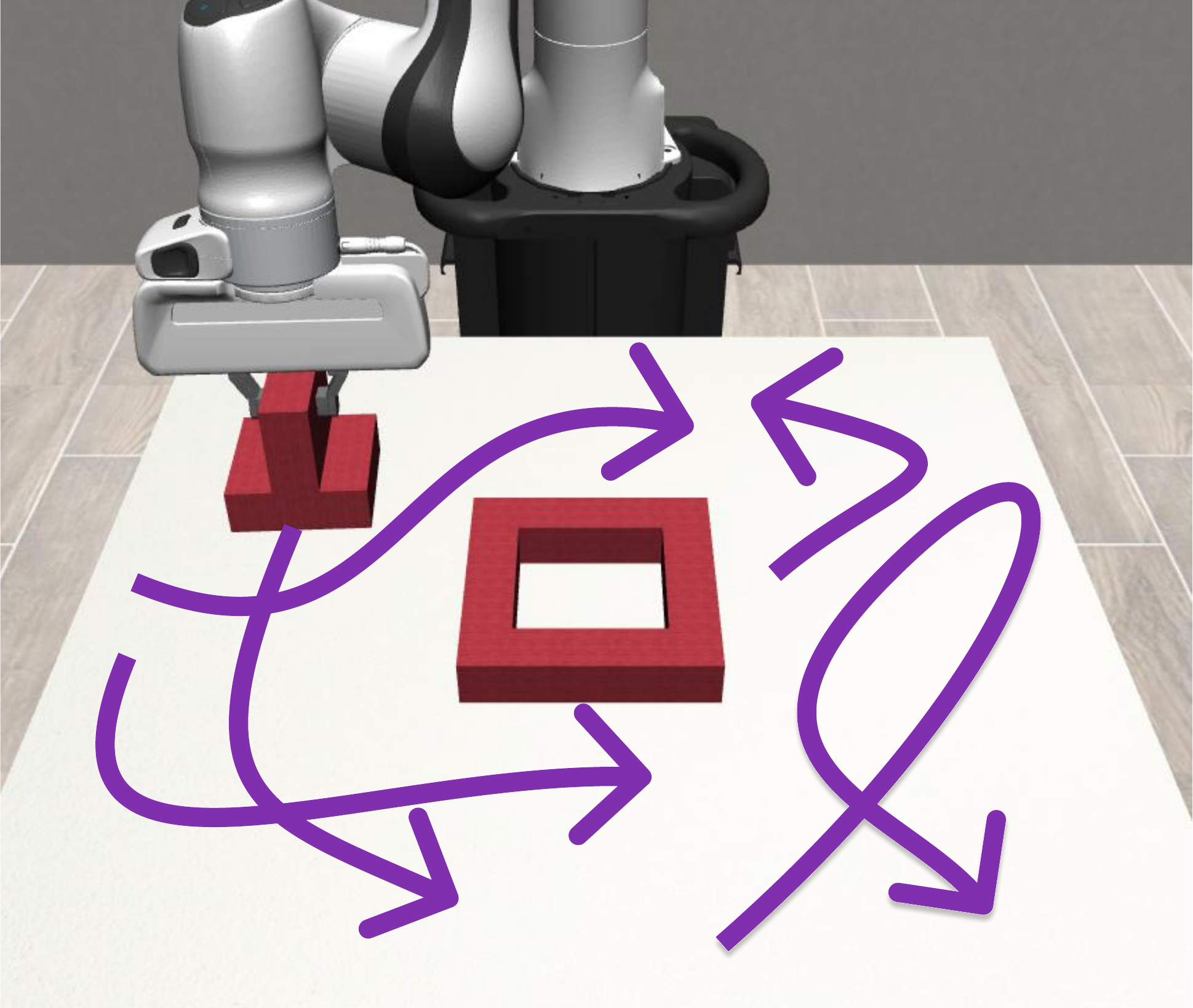}
        \end{subfigure} &
        \begin{subfigure}[b]{0.15\textwidth}
            \centering
            \includegraphics[width=\textwidth,height=2.0cm,keepaspectratio=false]{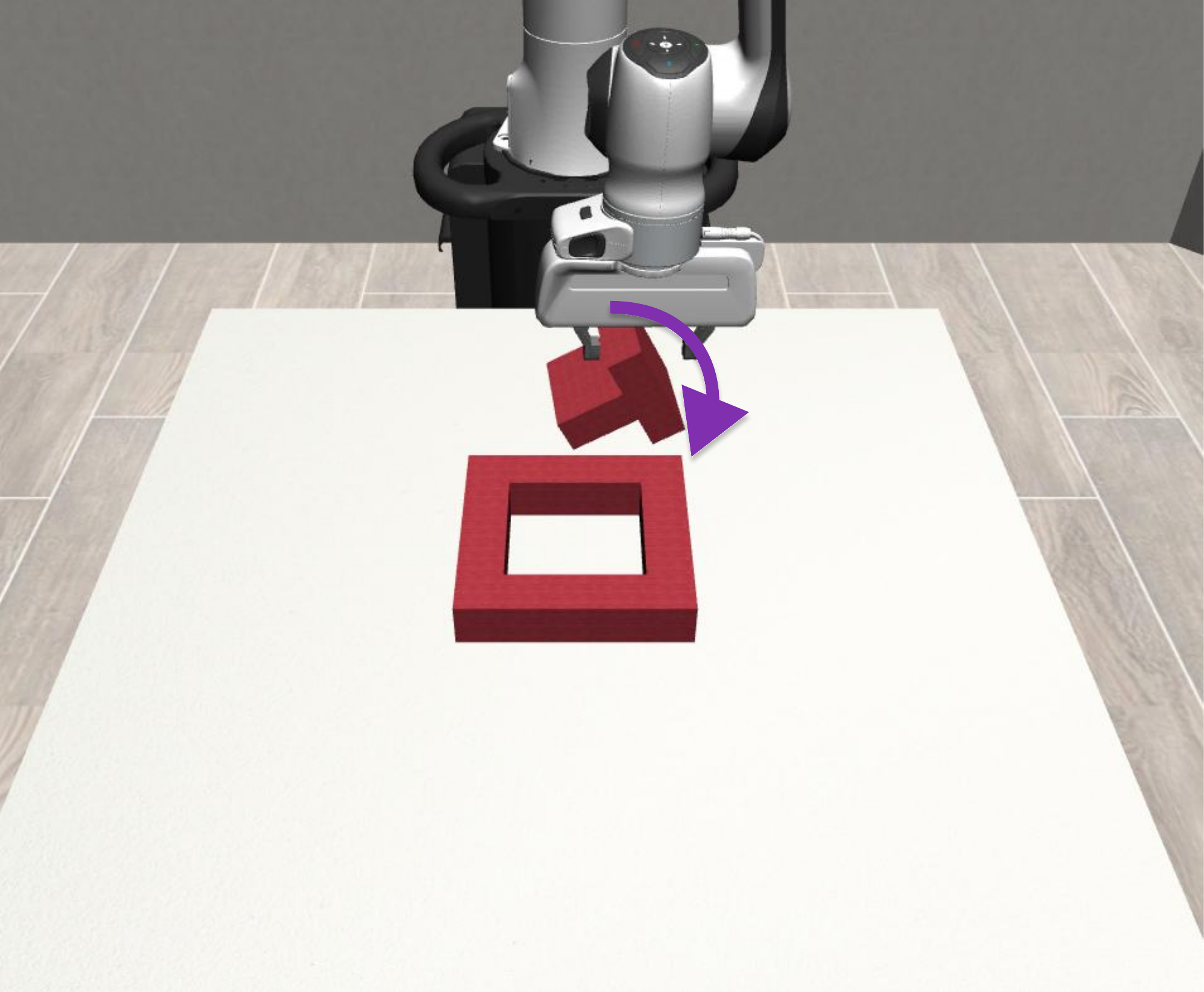}
        \end{subfigure} \\
        \begin{subfigure}[b]{0.15\textwidth}
            \centering
            \includegraphics[width=\textwidth,height=2.0cm,keepaspectratio=false]{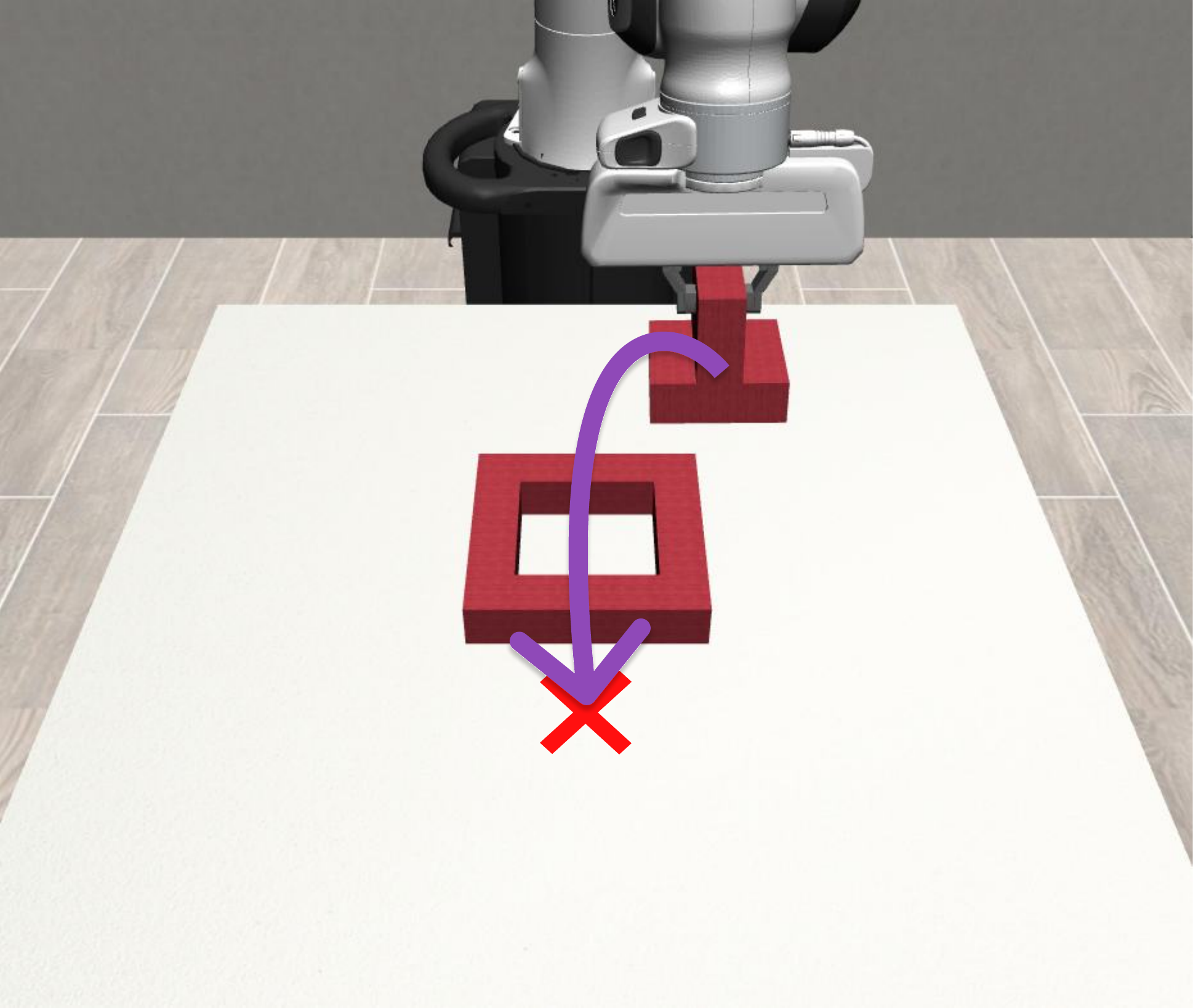}
        \end{subfigure} &
        \begin{subfigure}[b]{0.15\textwidth}
            \centering
        \includegraphics[width=\textwidth,height=2.0cm,keepaspectratio=false]{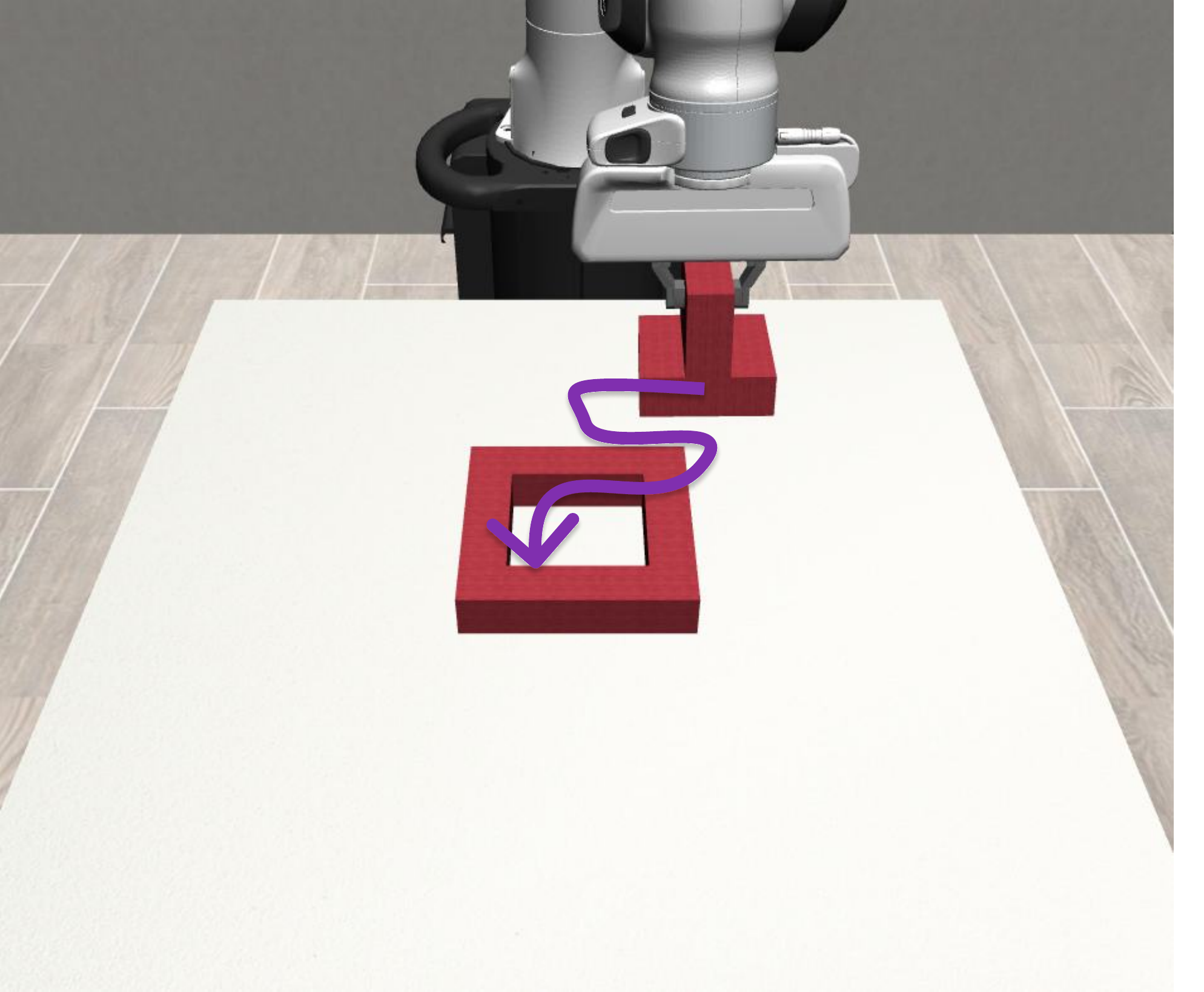}
        \end{subfigure}
    \end{tabular}
    \caption{\footnotesize{Visualizations of types of play data for the \twopiece task. (top left) high coverage play data --- the block is randomly picked up and moved around the table (top right) partial tipping --- the block is tipped over to its upright position (bottom left) failure --- the block is attempted to be inserted, but misses (bottom right) suboptimal --- the block is inserted into the hole, but in an inefficient manner}}
    \label{fig:2piece_vis}
\end{figure}

\textbf{\twopiece} This task involves picking up a T-shaped block and placing it into another square-shaped block. In the expert data, the block is initialized in a narrow region $p_0$ in the top right. In the ``high-coverage play" data, the block is initialized across the entire table $p_{test}$, and 
 randomly picked up and moved around. In the ``suboptimal" demonstrations, the block is also initialized in $p_0$, expert behavior is attempted, but is either executed poorly or unsuccessful (see Figure \ref{fig:2piece_vis}). In the tasks that include tipping, the block is initialized on its side, and must be re-oriented to be upright before being picked up. The ``partial tipping" non-expert data, the block is reoriented using a variety of tipping strategies. For the coverage experiment, we evaluate with the block initialized to $p_{test}$. For the suboptimal experiment, we evaluate with the block initialized from $p_0$.

 \textbf{\threading} This task involves threading a needle-like object into a small hole, requiring precision. The expert region $p_0$ is a narrow region on the right of the table. In the ``high-coverage play" data, the needle is initialized across the entire table $p_{test}$, and randomly picked up and moved around. 

\begin{figure}[!htb]
    \centering
    \setlength{\tabcolsep}{1pt} % Reduce space between columns
    \begin{tabular}{cc} % Use a tabular environment with two columns
        \begin{subfigure}[b]{0.2\textwidth}
            \centering
            \includegraphics[width=\textwidth,height=2.4cm,keepaspectratio=false]{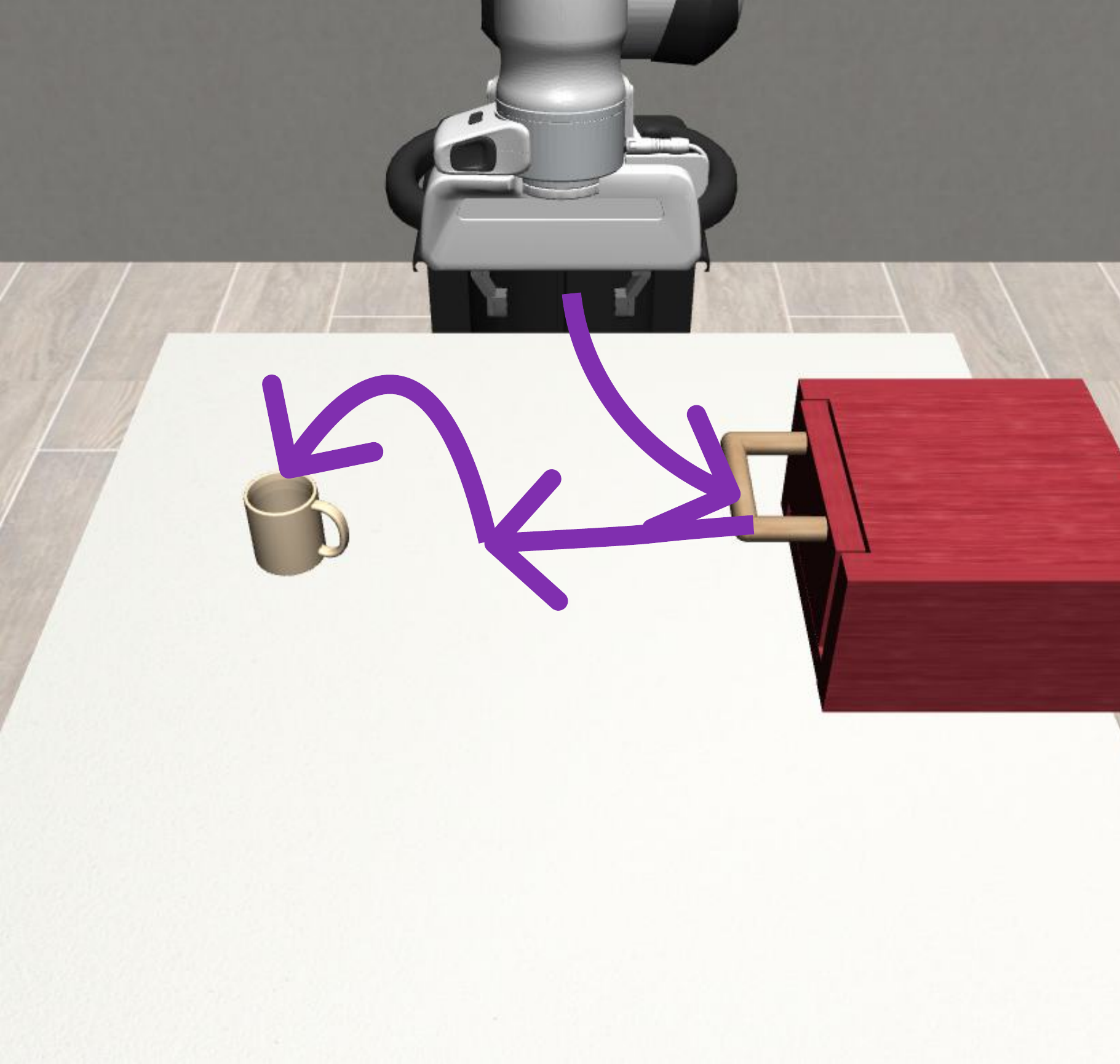}
            \caption{}
        \end{subfigure} &
        \begin{subfigure}[b]{0.2\textwidth}
            \centering
            \includegraphics[width=\textwidth,height=2.4cm,keepaspectratio=false]{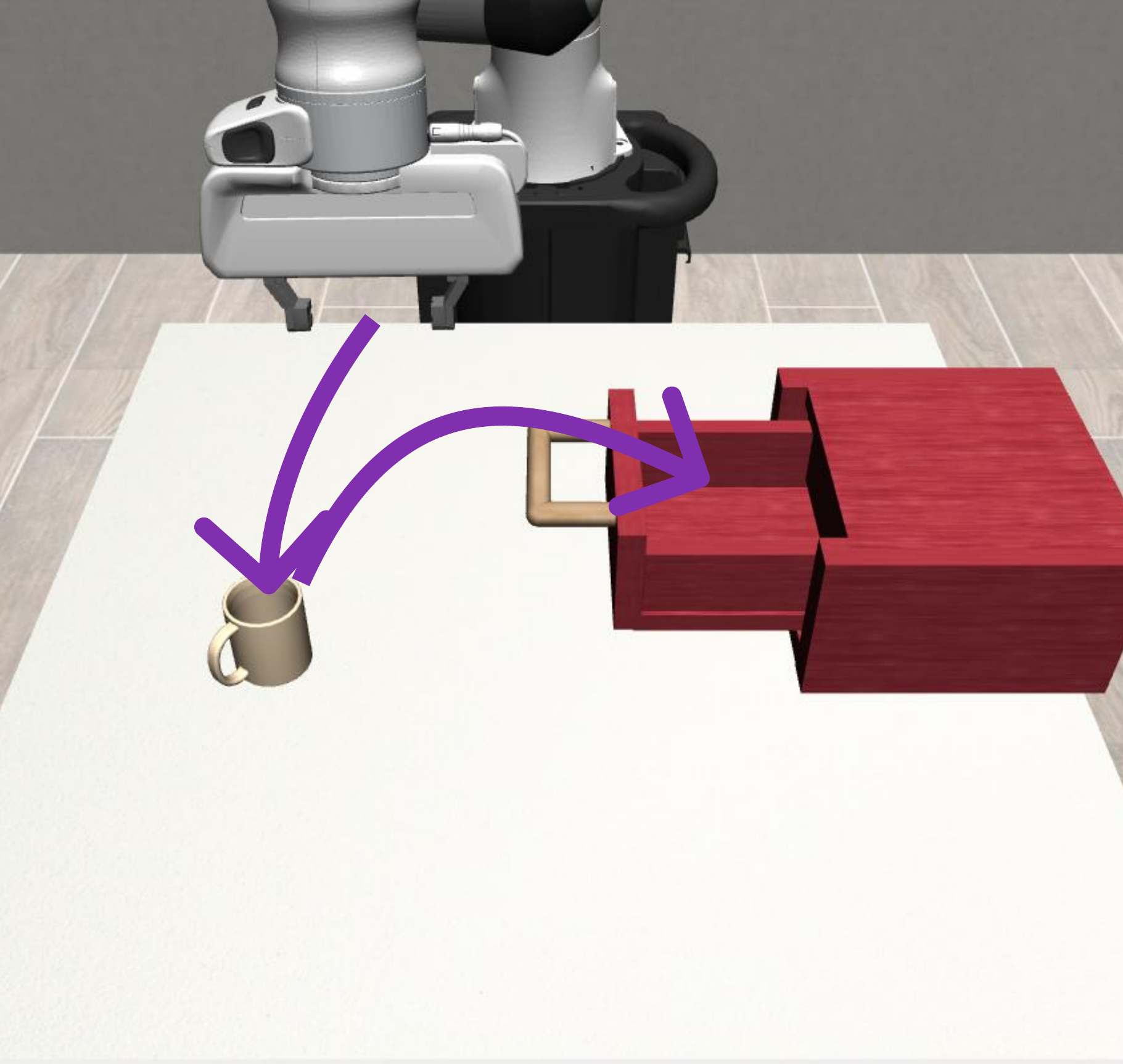}
            \caption{}
        \end{subfigure} 
    \end{tabular}
    \caption{\footnotesize{Visualization of non-expert data for the \mug task (a) drawer open only --- the demonstration only consists of opening the drawer and attempting to grasp the mug (b) place in drawer only --- the demonstration only consists of picking the mug up and placing it in the drawer}}
    \label{fig:mug_vis}
\end{figure}

 \textbf{\mug} This task involves opening a drawer, picking up a mug, and placing it into the drawer. Suboptimal data includes demonstrations where the demonstrator only opens the drawer, but then fails at grabbing the mug, and demonstrations where the drawer starts open, and the demonstrator places the mug in the drawer without demonstrating opening the drawer (see Figure \ref{fig:mug_vis}). In all cases, the mug is initialized from a small region $p_0$, which also serves as the evaluation region.

 \textbf{\lamp} This task involves picking up a lampshade and placing it on a partially assembled stand. The expert is collected from a narrow region $p_0$, while the ``high-coverage data" is collected from a wider initialization region $p_{test}$, where the lampshade is randomly pushed around the table. 

 \textbf{\oneleg} This task involves picking up a table leg and inserting it into a hole of a table. The expert is collected from a narrow region $p_0$. The suboptimal data consists of picking the leg from a wide range $p_{init}$, but failing to insert the leg into the hole successfully.

 \textbf{\cloth} This task involves folding a piece of cloth and then stacking it. The high coverage play involves randomly re-arranging and folding the cloth from various starting configurations. The expert data consists of picking up the cloth and stacking it from a narrow initial distribution where the cloth is already neatly placed.

\begin{table}[ht!b]
\centering
\tiny
\begin{tabular}{l|c|c}
\hline
Task & Dataset & Data Augmentation Threshold ($T$)\\
\hline
\hline
\squarenut (Coverage) & \makecell{Expert (200) \\ High Coverage Play (224)} & 0.15 \\
\hline
\squarehook (Coverage) & \makecell{Expert (175) \\ High Coverage Play (224)} & 0.15 \\
\hline
\twopiece (Coverage)& \makecell{Expert (200) \\ High Coverage Play (199)} 
& 0.08 \\
\hline
\twopiece (tipping) (Coverage) & \makecell{Expert (200) \\ High Coverage Play (199) \\ Partial Tipping (104)} & 0.08 \\
\hline
\threading (Coverage) & \makecell{Expert (200) \\ High Coverage Play (200)} & 0.08 \\
\hline
\hline
\mug (Suboptimal) & \makecell{Expert (20) \\ Suboptimal (85)} & 0.05 \\
\hline
\squarenut (Suboptimal) & \makecell{Expert (20) \\ Suboptimal (80)} & 0.0 \\
\hline
\twopiece (tipping) (Suboptimal) & \makecell{Expert (10) \\ Suboptimal (100)} & 0.0 \\
\hline
\hline
\lamp & \makecell{Expert (225) \\ High Coverage Play (276)} & 0.22 \\
\hline
\oneleg & \makecell{Expert (50) \\ Suboptimal (150) } & 0.1 \\
\hline
\cloth & \makecell{Expert (50) \\ High Coverage Play (50)} & 0.05 \\
\hline
\hline
\end{tabular}
    \caption{\footnotesize {Composition of datasets used for each task, along with the amount of data augmentation used. The number in parenthesis indicates the number of trajectories in that dataset.}}
    \label{tab:datasets}
\end{table}

\end{document}